\documentclass[twoside,11pt]{article}

\usepackage{jmlr2e}
\usepackage{appendix}
\usepackage{times}
\usepackage{graphicx} % more modern
\usepackage{subfigure}

\usepackage{algorithm}
\usepackage{algorithmic}

\usepackage{hyperref}

\usepackage{amsmath,amssymb}
\usepackage{wrapfig}
\usepackage{booktabs}
\newcommand{\SWITCH}[1]{\STATE \textbf{switch} (#1)}
\newcommand{\ENDSWITCH}{\STATE \textbf{end switch}}
\newcommand{\CASE}[1]{\STATE \textbf{case} #1\textbf{:} \begin{ALC@g}}
\newcommand{\ENDCASE}{\end{ALC@g}}

\newcommand{\DEFAULT}{\STATE \textbf{default:} \begin{ALC@g}}
\newcommand{\ENDDEFAULT}{\end{ALC@g}}
\newcommand{\DEFAULTLINE}[1]{\STATE \textbf{default:} }

\newenvironment{sketch}{{\bf Proof sketch:}\rm }{\hfill $\Box$ }

\newtheorem{thm}{Theorem}

\newtheorem{lem}{Lemma}

\usepackage{setspace}

\firstpageno{1}

\begin{document}

\title{Learning Modular Structures \\ from Network Data and Node Variables}

\author{\name Elham Azizi \email elham@bu.edu \\
       \name James E.\ Galagan \email jgalag@bu.edu \\
       \addr Departments of Biomedical Engineering and Microbiology\\
       Boston University \\
       Boston, MA 02215, USA
       \AND
       \name Edoardo M.\ Airoldi \email airoldi@fas.harvard.edu \\
       \addr Department of Statistics \\
       Harvard University \\
       Cambridge, MA 02138, USA
       %and The Broad Institute of Harvard and MIT\\
       }

%\editor{Leslie Pack Kaelbling}
%\editor{}

\maketitle

\begin{abstract}%   <- trailing '%' for backward compatibility of .sty file
A standard technique for understanding underlying dependency structures among a set of variables posits a shared conditional probability distribution for the variables measured on individuals within a group. This approach is often referred to as module networks, where individuals are represented by nodes in a network, groups are termed modules, and the focus is on estimating the network structure among modules. However, estimation solely from node-specific variables can lead to spurious dependencies, and unverifiable structural assumptions are often used for regularization.
Here, we propose an extended model that leverages direct observations about the network in addition to node-specific variables.
By integrating complementary data types, we avoid the need for structural assumptions. We illustrate theoretical and practical significance of the model and develop a reversible-jump MCMC learning procedure for learning modules and model parameters.
%
%inspired by module networks and stochastic blockmodels for learning structures from node variables (e.g. gene expression in biology or user activity in social networks) combined with network data (e.g. protein-DNA interactions in biology or followings/friendships in social networks).
%
We demonstrate the method accuracy in predicting modular structures from synthetic data and capability to learn influence structures in twitter data and regulatory modules in the {\it Mycobacterium tuberculosis} gene regulatory network.
\end{abstract}

\begin{keywords}
Module Networks, Blockmodels, Gene Regulatory Networks, ChIP-Seq, Reversible-Jump MCMC, Data Integration
\end{keywords}

\section{Introduction}
There is considerable interest in modeling dependency structures in a variety of applications. Examples include reconstructing regulatory relationships from gene expression data in gene networks or identifying influence structures from activity patterns such as purchases, posts, tweets, etc in social networks.
% BN
Common approaches for learning dependencies include using Bayesian networks and factor analysis \citep{Koller+Friedman:09}.% \citep{harman1976factoranalysis}.
%

%Regulatory interactions between genes drive various biological functions such as responses to environmental signals, metabolism and differentiation.
% In order to understand the mechanisms underlying cellular functions, we need to develop genome-scale models that decipher the immense complexity of regulatory interactions controlling gene expression.
%Identifying regulatory functions through biological experiments especially in a genome scale can be expensive, time-consuming and in some cases infeasible. Therefore, developing efficient computational models plays a key role in deciphering regulatory programs and guiding experiments.
%There is considerable interest in modeling regulatory relationships based on gene expression.
% and factor analysis \citepp{harman1976factoranalysis}.
% Why Module

%Modular behavior in regulatory networks has been shown to natural and interpretable
%ihmels2004defining
%In other domains, e.g. in social networks, communities with similar interests or affiliations may have similar behavior in communicating messages in response to news-outbreaks or similar purchases in response to marketing advertisements \citepp{kozinets1999tribalized,aral2009distinguishing}. Alternatively, in stock markets, sectors of stock prices respond together to certain driving factors.

% MN
% MN
Module networks \citep{Segal2005,Segal2003} have been widely used to find structures (e.g. gene regulation) between groups of nodes denoted as modules, based on measurements of node-specific variables in a network (e.g. gene expression).
%
%A widely used method for finding regulatory structures is module networks \citep{Segal2005,Segal2003}, which finds regulatory structures between groups of genes.
The motivation lies in that nodes that are influenced or regulated by the same parent node(s), have the same conditional probabilities for their variables. For example, in gene regulatory networks, groups of genes respond in concert under certain environmental conditions \citep{qi2006modularity} and are thus likely to be regulated by the same mechanism. In other domains, such as social networks, communities with similar interests or affiliations may have similar activity in communicating messages in response to news-outbreaks or similar purchases in response to marketing advertisements \citep{kozinets1999tribalized,aral2009distinguishing}.
%it is informative to identify regulatory mechanisms that drive groups or modules of functionally-related genes within the network.
%
%Thus, the assumption is that co-regulated genes, represented as a module, have similar probability distributions for their expression conditioned on their shared regulators.
%
%, i.e. $Mj \bigcap Pa_{k} \neq \emptyset$.
%

% problem
However, inferring dependencies merely from node-specific variables can lead to higher rate of false positives \citep{michoel2007validating}. For example, a dependency might be inferred between two unrelated nodes due to existing confounding variables. This can introduce arbitrary or too many parents for a module.
%Moreover, detection of dependencies are restricted to observation of correlation between variables (false negatives)\HAS{why these are bad?}.
%To avoid over-fitting, additional structural assumptions such as maximum number of modules or maximum number of parents per module may be required for utilizing this method, which present additional inductive bias.
%
% Why Module
%It can be beneficial to identify groups or modules within these interaction networks. Modular behavior can be natural and interpretable in some domains such as gene regulatory networks, which consist of partitions of genes acting in concert
%under certain environmental cues \cite{qi2006modularity}.%ihmels2004defining
%
% problem
%However, identifying regulatory links based on co-expression presents certain limitations. For example, a link can be inferred between two genes that are not directly connected, but show correlated expression and thus lead to false positive predictions~\citep{michoel2007validating}.
%Moreover, detection of dependencies are restricted to observation of correlation between variables (false negatives)
To avoid over-fitting in inferring module networks, additional structural assumptions such as setting the maximum number of modules or maximum number of parents per module may be required. This in turn presents additional inductive bias and results become sensitive to assumptions. Moreover, searching through the entire set of candidate parents for each module is computationally infeasible.

\begin{figure*}[width=0.9\textwidth]
 % \flushleft
 \centering
  % Requires \usepackage{graphicx}
  \includegraphics[width=1\textwidth]{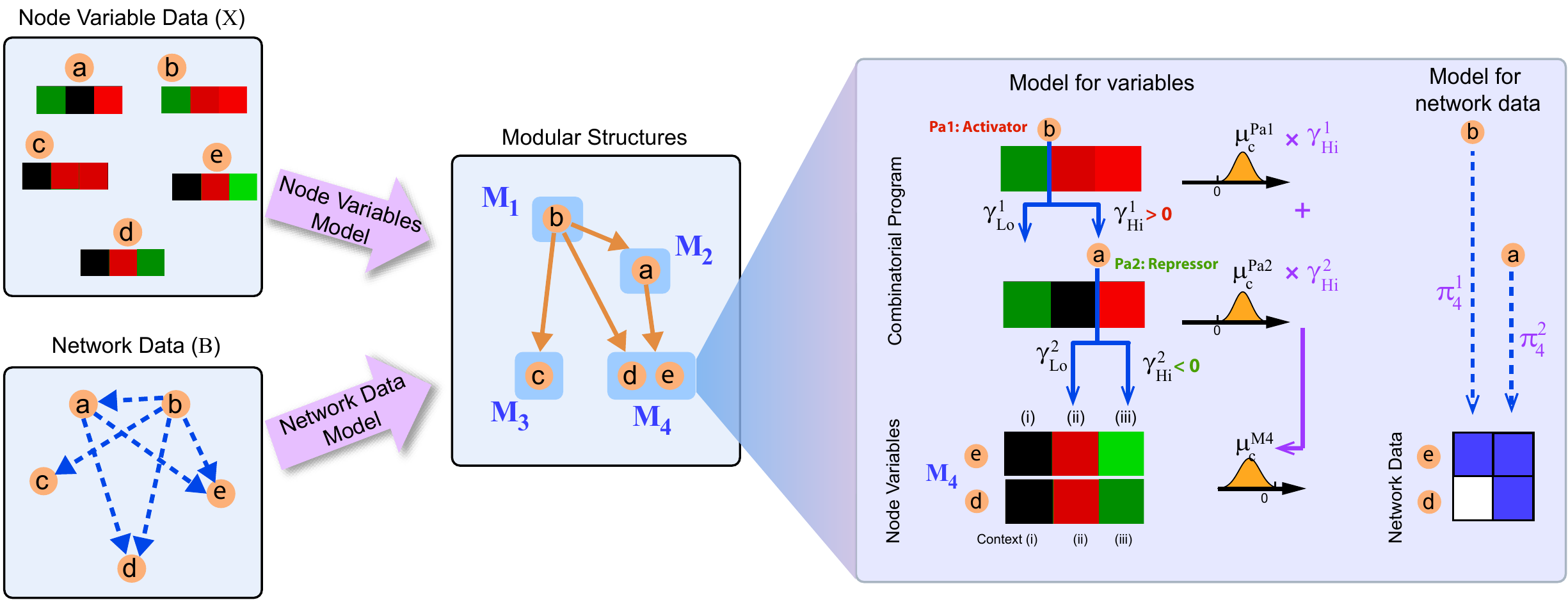}\\
  \caption{{\bf Illustration of proposed model:} Modular structures are learned from node variables (e.g. gene expression) and network data (e.g. protein-DNA interactions). Node variables are color-coded ranging from green (low) to red (high). A number of parents are assigned to each module (orange links). A combinatorial program is inferred for each module; example shown for module $M_4$. }
  \label{fig_mn}
  %\vspace{-3mm}
%\end{wrapfigure}
\end{figure*}
Alternatively, we can take advantage of existing network data and by integrating node interactions with node variables, we can avoid structural assumptions. For example, to learn gene regulatory networks, we can use protein-DNA interaction data, which shows physical interactions between proteins of genes (known as Transcription Factors) with promoter regions of other genes, leading to regulation of transcription (and expression) of the latter genes.
This data can be measured using chromatin immunoprecipitation of DNA-bound proteins, i.e. ChIP-ChIP or ChIP-Seq technologies, which have shown to be informative of regulation \citep{Galagan2013,liu2013bayesnetchipseq,celniker2009unlocking,yeang2004physical}.
As another example, to learn influence structures in a twitter network, we can integrate the network of who-follows-who with measurements of users activities.
%These datasets identified dense interconnections between regulators themselves, suggesting possibilities of combinatorial regulations. Imposing modular structures on interaction networks can reveal systems-level regulatory programs and architecture.

Identifying modules or block structures from network data has been well-studied, e.g., using stochastic blockmodels  \citep{wang1987stochastic,snijders1997blockmodel,airoldi2008mixed,Airoldi_Costa_Chan_2013} in the area of social network modeling \citep{Goldenberg_Zheng_Fienberg_2009,azarigraphlet,Choi_Wolfe_Airoldi_2012}. Stochastic blockmodels assume that nodes of a network are members of latent blocks, and describe their interactions with other nodes with a parametric model. However, models for inferring modular structures from both data node variables and network data are relatively unexplored and of interest in many applications.
%
%These datasets identified dense interconnections between regulators themselves, suggesting possibilities of combinatorial regulations. Imposing modular structures on interaction networks can reveal systems-level regulatory programs and architecture.
%
\subsection{Contributions}
In this paper, we propose an integrated probabilistic model inspired by module networks and stochastic blockmodels, to learn dependency structures from the combination of network data and node variables.
We consider network data in terms of directed edges (interactions) and model network data using stochastic blockmodels. Intuitively, by incorporating complementary data types, a node which is likely to have directed edges to members of a module as well as correlation with variables of module will be assigned as parent.
%relational data as well as correlat regulatory function to those Transcription Factors (TFs), that have both physical interaction with gene promoters and predictive power in explaining their expression.
% solution
%Incorporating complementary relational data, can improve accuracy by avoiding false assignments of indirect regulators or regulators with correlated variables \citep{de2010advantages}.
 The use of network data enhances computational tractability and scalability of the method by restricting the space of possible dependency structures. We also show theoretically that the integration of network data leads to model identifiability, whereas node variables alone can not.
%
%XXX\E{We only consider non-overlapping modules while the mmsb main idea is mixed memberships for variables, how to relate?...}
%gene expression data with ChIP-Seq interaction data that capture
%direct bindings as opposed to indirect regulation (or mere co-expression). We would like to model combinatorial interactions between directly regulating TFs that explain different gene expressions in different conditions. \\

%Examples in social networks can include integrating number of posts on facebook (as object variables) with number of messages sent between friends (relational data) to identify structures of influence.
%
Our model captures two types of relationships between variables of modules and their parents, including small changes of variables due to global dependency structure and condition-specific large effects on variables based on parent activities in each condition. Based on these relationships, we infer a combinatorial program \citep{yeang2006modeling,Segal2003} for each module, showing how multiple parents interact in regulating the module.

%We present a Reversible-Jump MCMC inference which provides posterior distributions of parameters including the number of modules and parents, which can be interpreted in the context of data.
%
%In modeling expression, we assign different parameters for each context in each module. Each context is assigned to one combination of regulator states (ON/OFF) in a module. Therefore, the parameters of conditional probability distributions of expressions in a bi-cluster of module-context are the same. The expression of genes are then represented as a mixture of distributions for each context, based on the expressions of the gene regulators and the structure of the dependency tree for the module which it belongs to.
For estimation of parameters, we use a Gibbs sampler instead of the deterministic algorithm employed by Segal et al. to overcome some of the problems regarding multi-modality of model likelihood \citep{Joshi2009}.
%  A deterministic optimization algorithm is used by Segal et al. to search simultaneously for a partition of genes into modules and regulatory structures for each module, which involves searching over multiple local optima .
 We also solve the problem of sensitivity to choice of maximum number of modules using a reversible-jump MCMC method which infers the number of modules and parents based on data. The probabilistic framework infers posterior distributions of assignments of nodes to modules and thus does not face restrictions of non-overlapping modules \citep{airoldi2008mixed,Airoldi:2013fk}.
\subsection{Related Work}
Other works have also proposed integrating different data types, mostly as prior information, for improvement in learning structures \citep{Werhli2007,Imoto2003a,mitra2013integrative}.
Assumptions such as sparse priors have been used in other works to improve modeling of network interactions between groups of nodes \citep{yan2012sparse}. Our approach is different in that we consider additional data types also as observations from a model of dependency structures.
Our model thus considers both network edges and node variables as data observed from the same underlying structure, providing more flexibility for the model.
Moreover, we utilize data integration to identify structures between groups of nodes (modules) as opposed to individual nodes.
Despite the similarity in the framework of our model to module networks, our model for variables has differences in relating modules to their parents, giving more accurate and interpretable dependencies. Also, the integration of network data is novel.
 %We also show theoretically that integrating interaction data can lead to identfiability of the model, whereas expression data alone cannot.
%
%
Regarding the learning procedure, prior work has been done on improving module network inference by using a Gibbs sampling approach \citep{Joshi2009}. We take a step further and use a reversible-jump MCMC procedure to learn the number of modules and parents from data as well as parameter posteriors.
%\footnote{There are alternative methods to RJMCMC, such as the saturation method~\citep{brooks2003efficient}, however, considering the combinatorial nature of the modules in this problem, keeping track of former sampled modules is infeasible.}.
Our method can also allow restricting the number of modules based on context, with a narrow prior. By adjusting this prior, we have multi-resolution module detection.
%\section{Approach}
%
%Our approach for learning .. is summarized in figure ..
%Equation~(\ref{eq:01}) Text Text Text Text Text Text  Text Text Text Text Text Text Text Text Text  Text Text Text Text Text Text. Figure \ref{fig:02} shows that the above method  Text Text Text Text  Text Text Text Text Text Text  Text Text.  \citepalp{Boffelli03} might want to know about  text text text text ……
%
%
%\vspace{-4mm}
%\begin{methods}
%\section{Model}
%
\section{Model of Modular Structures}
In the framework of module networks, dependencies are learned from profiles of node variables (e.g. gene expressions) for each node (e.g. gene), as random variables $\{X_1, ..., X_N\}$. The idea is that a group of nodes with common parents (e.g. co-regulated genes) are represented as a module and have similar probability distributions for their variables conditioned on their shared parents (regulators). Figure \ref{fig_mn} shows a toy example where node variable data are shown in green-to-red heatmaps and network data with dashed arrows \citep{Airoldi2007}.
%
%\begin{wrapfigure}{r}{4.5in}
%
%
%\vspace{-0.05in}
A module assignment function $\mathcal{A}$ maps nodes $\{1,...,N\}$ to $K$ non-overlapping modules. A dependency structure function $\mathcal{S}$ assigns a set of parents $Pa_{j}$ from $\{1,...,R\}$ known candidate parents (possible regulators/influencers), which are a subset of the $N$ nodes, to module $M_j$ (figure \ref{fig_mn}). In the toy example, nodes $d,e$ are assigned to the same module $M_4$ and $b,a$ are assigned as their parents.
In cases where multiple parents drive a module, e.g. $a,b$ affecting $M_4$, combinatorial effects are represented as a decision tree (regulatory program) and each combination of parents activities, defined as a context, is assigned to a cluster of conditions (experiments). In figure \ref{fig_mn}, parent $b$ has an activating effect while $a$ represses $M_4$, hence, $e,d$ are active in context $(ii)$ where only $b$ is active and $a$ is not. Inferring this decision tree in the context of different applications shows how multiple parents act together in influencing a group of nodes, e.g. in a gene network, multiple transcription-factor (TF) proteins act as regulators together to express a group of genes.

Given this framework, our model considers variables and network data as two types of observation from the same underlying modular structure. This structure is encoded based on assignments to modules ($\mathcal{A}$) and parents for each module ($\mathcal{S}$). In the example of gene networks, in each module, TF-gene interactions are likely to be observed between TFs and upstream regions of genes in the module while combinations of expressions of TFs explain expressions of genes.
%This effect is modeled in two types of global and condition-specific components.
%\subsection*{Model} \label{sec_model}
%\vspace{-0.1in}
%
%\section{Model} \label{sec_model}
%\vspace{-0.1in}
%
%\begin{wrapfigure}{r}{4.5in}
%
%
%In the case of multiple parents for a module, combinatorial interactions \cite{Buchler:2003:Proc-Natl-Acad-Sci-U-S-A:12702751} can occur, represented as a regression tree in which clusters of samples (or conditions) are assigned to one context. Clustering samples or conditions in addition to variables can guide experimental design for validation of regulations \cite{de2010advantages}.
%
%We present improvements to the module net model and inference and integrate directed interaction data as an observation (additional data) generated from the same generative process.
%The framework presented A ChIP-Seq Binding event between regulator $r \in \{1,...,R\}$ and gene $n\in \{1,...,N\}$ is defined as $B_{r\rightarrow n}$
%
%\begin{figure}
%  \centering
%  % Requires \usepackage{graphicx}
%  \includegraphics[width=.94\textwidth]{model_fig5.pdf}\\
%  \caption{{\bf (A)} Example module network {\bf (B)} A combinatorial regulatory program is inferred for each module; example shown for $M_4$. }
%  \label{fig_mn}
%  \vspace{-5mm}
%%\end{wrapfigure}
%\end{figure}
%
%\vspace{-0.1in}
\subsection{Modeling Node Variables}
%\vspace{-0.05in}
We model variables for nodes $\{1,...,N\}$ in each condition or sample $c \in {1,...,C}$ with a multivariate normal represented as $ {\bf X_c} \sim \mathcal{N} ({\boldsymbol \mu_c}, \Sigma)$,
%
%\begin{equation}
%    {\bf X_c} \sim \mathcal{N} ({\boldsymbol \mu_c}, \Sigma)
%\end{equation}
%
where $\bf{X_c}$ is a $N \times 1$ vector, with $N$ being the total number of nodes. The covariance and mean capture two different aspects of the model regarding global dependency structures and context-specific effects of parents, respectively, as described below.

%%%% SIGMA
We define the covariance ${\bf \Sigma}$ to be independent of conditions and representing the strength of potential effects of one variable upon another, if the former is assigned as a parent of the module containing the latter. In the example of gene expressions, $\Sigma$ may represent the affinity of a Transcription-Factor protein to a target gene promoter.
The modular dependencies between variables imposes a structure on ${\bf \Sigma}$. To construct this structure, we relate node variables to their parents through a regression ${\bf X_c} = W{\bf X_c} + {\boldsymbol \epsilon}$
where ${\boldsymbol \epsilon} = \mathcal{N}({\bf m_c},I)$. $W$ is a $N \times N$ sparse matrix in which element $W_{nr}$ is nonzero if variable $r$ is assigned as a parent of the module containing variable $n$. Here we assume $W_{nr}$ has the same value for $\forall n \in M_k, \forall r \in Pa_{k}$, which leads to identifiability of model (as explained in section \ref{theory}.
Then, assuming $I-W$ is invertible, ${\bf X_c} = (I-W)^{-1} {\boldsymbol \epsilon}$  which implies ${\bf \Sigma}=(I-W)^{-T}(I-W)^{-1}$.
%
%\begin{wrapfigure}{}{0.4\textwidth}
%  %\centering
%  % Requires \usepackage{graphicx}
%  \includegraphics[width=0.4\textwidth]{genproc11.pdf}
%  \vspace{-4mm}
%  \caption{Graphical representation of model}\label{model}
%  \vspace{-2mm}
%\end{wrapfigure}
%\begin{equation}\label{sigma}
%{\bf \Sigma}=(I-W)^{-T}(I-W)^{-1}
%\end{equation}
%
%\begin{equation}
%    X - WX = \epsilon
%\end{equation}
%\begin{equation}
%    X(I-W) = \epsilon
%\end{equation}
%
%
%which we model as
%\begin{equation}
% X = N(\bf{\mu}, \bf{\Sigma}
%\end{equation}
Therefore, we impose the modular dependency structure over $\Sigma$ through $W$, which is easier to interpret based on $\mathcal{A,S}$ assignments.
%
%We will learn $\mu$ such that the elements for one module are the same. In general the $\mu$ and $A$ will be the same for every bi-cluster (a context of a module).
%
%%%%   MEAN

We define variable means $\boldsymbol{\mu}_c$, based on parents as described below. First, based on the modular structure of nodes, we can partition the mean vector as ${\boldsymbol \mu_c} = [{\boldsymbol \mu_{c}^1} ... {\boldsymbol \mu_{c}^K}]^T$,
%
%\begin{equation}
%{\boldsymbol \mu_c} = [{\boldsymbol \mu_{c}^1} ... {\boldsymbol \mu_{c}^K}]^T
%\end{equation}
%
where each $\boldsymbol{\mu}_{c}^k$ for $k=1,...,K$ is a $1\times N_k$ vector with $N_k$ equal to the number of nodes in module $k$.
In modules where there is more than one parent assigned, combinations of different activities of parents, creating a context, can lead to different effects. The binary state of parent $r \in Pa_{k}$ is defined by comparing its mean to a split-point $z_k^r$, corresponding to a mixture coefficient for that state $\gamma_{Lo}^r$ or $\gamma_{Hi}^r$, as: $\gamma_c^{r} = \gamma_{Lo}^r H(z_k^r - \mu_c^r) + \gamma_{Hi}^r H(\mu_c^r - z_k^r)$, where $H(\cdot)$ is a unit step function.
%
%\

\begin{figure}%{0.5\textwidth}
  \centering
  %\vspace{-16mm}
  %\vspace{-3mm}
  % Requires \usepackage{graphicx}
  \includegraphics[width=0.5\textwidth]{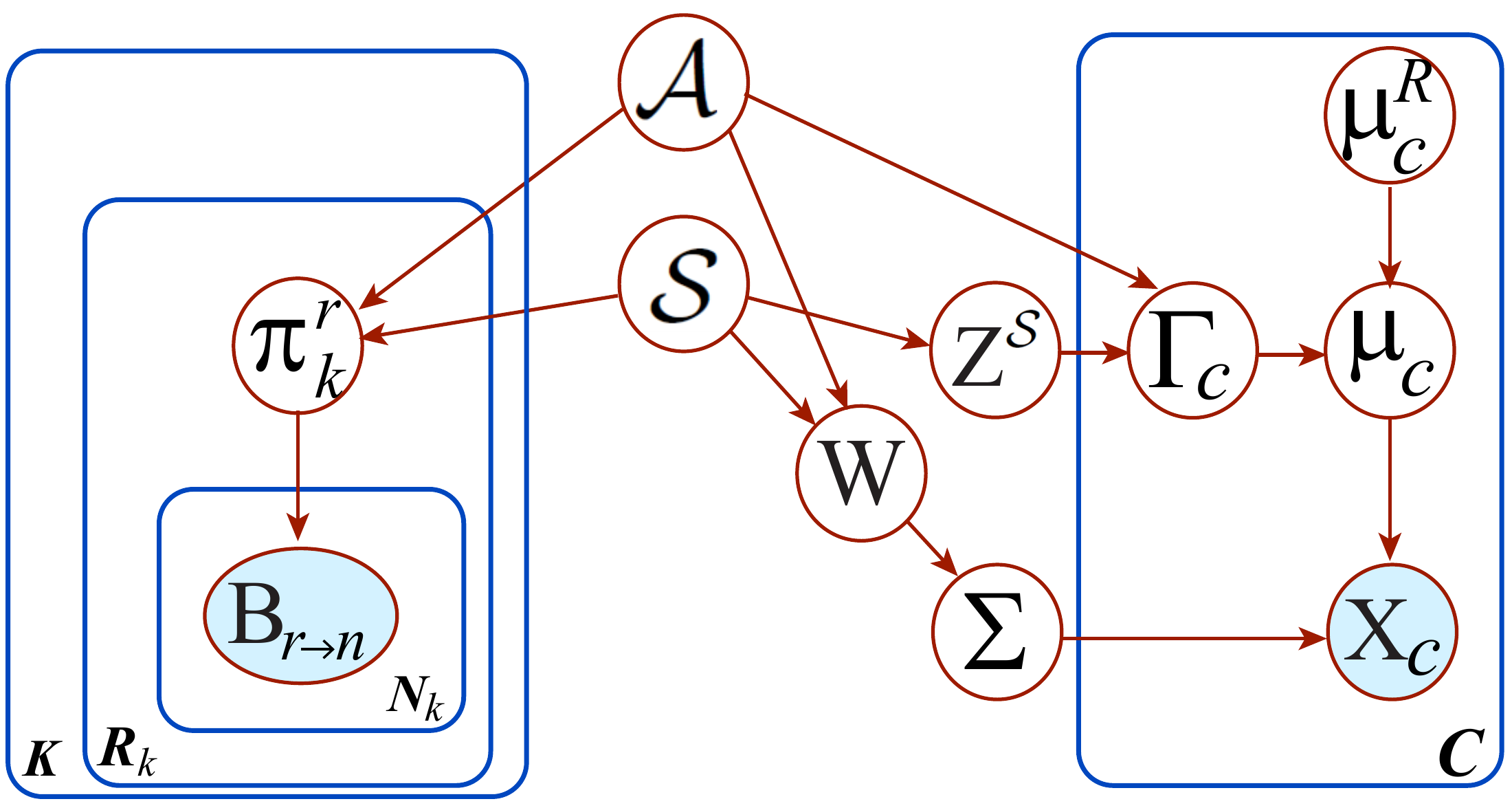}
  %\vspace{-4mm}
  %\vspace{-3mm}
  \caption{{\bf Graphical representation of model:} The assignments of nodes to modules $\mathcal{A}$ and parents for modules $\mathcal{S}$ represent modular dependency structures, from which we observe node variables $\mathbf{X_c}$ in each condition $c$ and network data $B_{r  \rightarrow n}$ between a parent $r$ and a node $n$. Means of node variables $\mu_c$ are determined from parent means $\mu_c^R$ with mixing coefficients $\Gamma$ determined based on parent split-points $Z$.}\label{model}
 % \vspace{-6mm}
\end{figure}

The combination of different activities are represented as a decision tree for each module $k$ (figure \ref{fig_mn}).
We represent a context-specific program as dependencies of variable means on parents activities in each context, such that $\boldsymbol{\mu}_{c}^k$ for module $k$ is a linear mixture of means for parents of that module:
%
%\vspace{-0.1in}
%\begin{equation}
$\boldsymbol{\mu}_{c}^k  =  \sum_{r=1}^{R_k} {\boldsymbol \gamma_{c}^r} {\boldsymbol \mu_{c}^{Pa_k}}$
%\end{equation}
%\vspace{-0.02in}
%
where $R_k$ is the number of parents $Pa_{k}$ and ${\boldsymbol \gamma_{c}^r}$ are similar for all conditions $c$ occurring in the same context.
Thus, in general we can write ${\boldsymbol \mu_c}=  {\Gamma}_c {\boldsymbol \mu_{c}^R}$,
%
%\begin{equation}\label{mu_c}
%{\boldsymbol \mu_c}=  {\Gamma}_c {\boldsymbol \mu_{c}^R}
%\end{equation}
%
where ${\boldsymbol \mu_{c}^R}$ contains the means of parents $1,...,R$ in condition $c$. The $N \times R$  matrix ${\Gamma}_c$ has identical rows for all variables in one module based on the assignment functions $\mathcal{A,S}$. The graphical model is summarized in figure \ref{model}. Thus the model for object variables would be:
%
%\vspace{-0.02in}
%\begin{equation}
 $   {\bf X_c} \sim \mathcal{N}({\Gamma}_c {\boldsymbol \mu_{c}^R}, (I-W)^{-T}(I-W)^{-1})$.
%\end{equation}
%\vspace{-0.02in}
%Furthermore, based on the regulatory program $\mathcal{T}_k$ of module $k$, and the values in another condition $c'$ which belongs to the same context as $c$ of the same module, would also be similar).
%
%
%
%
%

%%%% LIKELIHOOD
Given independent conditions, the probability of data $\mathbf{X}=\mathbf{[X_{1},...,X_{C}}]$ for $C$ conditions given parameters can be written as multiplication of multivariate normal distributions for each condition: $P(\mathbf{X}|\mathcal{A},\mathcal{S},\Theta,\Sigma,Z^S) = \prod_{c=1}^C P(\mathbf{X_{c}}|\mathcal{A},\mathcal{S},\theta_c,\Sigma,Z^S)$,
%
%\vspace{-0.05in}
%\begin{equation}\label{}
%  P(\mathbf{X}|\mathcal{A},\mathcal{S},\Theta,\Sigma,Z^S) = \prod_{c=1}^C P(\mathbf{X_{c}}|\mathcal{A},\mathcal{S},\theta_c,\Sigma,Z^S)
%\end{equation}
%\vspace{-0.02in}
%
where $\Theta=\{\theta_1,...,\theta_C\}$ denotes the set of condition-specific parameters $\theta_c=\{ {\boldsymbol \mu_{c}^R},\Gamma_c \}$ for $c=1,...,C$ and $Z^S$ denotes the set of parent split-points for all modules. Then for each condition we have: $P(\mathbf{X_c}|\mathcal{A},\mathcal{S},\theta_c,\Sigma,Z^S) =\frac{1}{(2\pi)^{N/2}|\mathbf{ \Sigma}^{1/2}|}exp(-\frac{1}{2}( \mathbf{X_c}-{\boldsymbol \mu_c})^T \mathbf{ \Sigma}^{-1}(\mathbf{X_c}-{\boldsymbol \mu_c}))$.
%
%\vspace{-0.05in}
%\begin{equation}
%%\begin{equation}\label{multi}
%P(\mathbf{X_c}|\mathcal{A},\mathcal{S},\theta_c,\Sigma,Z^S) =
%%\end{equation}
%%\begin{equation}
%\frac{1}{(2\pi)^{N/2}|\mathbf{ \Sigma}^{1/2}|}exp(-\frac{1}{2}( \mathbf{X_c}-{\boldsymbol \mu_c})^T \mathbf{ \Sigma}^{-1}(\mathbf{X_c}-{\boldsymbol \mu_c}))
%\end{equation}
%%\end{align*}

Hence, this model provides interpretations for two types of influences of parents. By relating the distribution mean for variables in each module and in each condition to means of their assigned parents (figure \ref{fig_mn}.B), we model condition-specific effects of parents. Based on the states of parents in different contexts (partitions of conditions), this leads to a bias or large signal variations in node variables. Whereas, small signal changes (linear term) are modeled through the covariance matrix $\Sigma$ which is independent of condition and is only affected by the global wiring imposed by dependency structures.
\subsection{Modeling Network Data}
%\vspace{-0.05in}
 Network data, as a directed edge between a parent $r \in \{1,...,R\}$ and node $n \in M_k$, when $r$ is assigned as a parent of the module $r \in {Pa_{k}}$ is defined as a directed link $B_{r\rightarrow n}$ where
%
%\vspace{-0.02in}
\begin{equation}\label{pi_bernoulli1}
    P(B_{r \in Pa_{k} \rightarrow n\in M_k}|\mathcal{A},\mathcal{S},\pi_{k}^r) \sim Bernoulli(\pi_{k}^r)
\end{equation}
The parameter $\pi_{k}^r$ defines the probability of parent $r$ influencing module $M_k$ (figure \ref{model}). In the gene network example, an interaction between a Transcription Factor protein binding to a motif sequence, upstream of target genes, which is common in all genes of a module can be observed using ChIP data. Therefore, directed interactions from parents to all nodes in a module would be $P(B_{M_k}| \mathcal{A},{\mathcal S},\boldsymbol{\pi}_{k}) =  \prod_{r\in Pa_{k}} \prod_{n\in M_k} P(B_{r  \rightarrow n}|\mathcal{A},\mathcal{S},\pi_{k}^r)$,
%
%\vspace{-0.05in}
%\begin{equation}\label{pi_bernoulli2}
%    P(B_{M_k}| \mathcal{A},{\mathcal S},\boldsymbol{\pi}_{k}) =  \prod_{r\in Pa_{k}} \prod_{n\in M_k} P(B_{r  \rightarrow n}|\mathcal{A},\mathcal{S},\pi_{k}^r)
%\end{equation}
%\vspace{-0.05in}
%
where $\boldsymbol{\pi_{k}}$ is the vector of $\pi_{k}^r$ for all $r \in Pa_{k}$ and for all nodes we have:
%
%\vspace{-0.1in}
\begin{align}\label{Ball}
%\vspace{-0.05in}
    &P(\mathbf{B}| {\mathcal A,S},\boldsymbol{\pi}) =  \prod_{k=1}^K \prod_{r\in Pa_{k}} \prod_{n\in M_k}  P(B_{r  \rightarrow n}|\mathcal{A},\mathcal{S},\pi_{k}^r) \nonumber\\
    &= \prod_{k=1}^K \prod_{r\in Pa_{k}} (\pi_{k}^r)^{s_{rk}} (1- \pi_{k}^r)^{|M_k| - s_{rk}} \nonumber \\
     &\prod_{r' \not\in Pa_{k}} (\pi_{0})^{s_{r'k}} (1- \pi_{0})^{|M_k|- s_{r'k}}
    % \vspace{-0.1in}
\end{align}
with $\boldsymbol{\pi}=\{ \boldsymbol{\pi}_{1},...,\boldsymbol{\pi}_{K} \}$ and $s_{rk} = \sum_{n\in M_k} (B_{r \rightarrow n})$ is the sufficient statistic for the network data model and $|M_k|$ is the number of nodes in module $k$ and $\pi_0$ is the probability that any non-parent can have interaction with a module. In gene regulatory networks, $\pi_0$ can be interpreted as basal level of physical binding that may not necessarily affect gene transcription and thus regulate a gene.

In the context of stochastic blockmodels, the group of parents assigned to each module can be considered as an individual block and thus our model can represented as overlapping blocks of nodes.
%
%Not all observed pairwise interactions can be interpreted as regulation or influence, and we represent these non-functional (silent) factors with ${\mathcal L}$. For example, Transcription Factors may show physical binding to gene promoters while their expression does not show correlation with the target genes. It has been shown that only a portion of physical protein-DNA interactions have regulatory roles XX(ref).
%
%\subsection{Joint Model Posterior}
%
%\vspace{-0.05in}

The likelihood of the model $\mathcal{M}=\{ \mathcal{ A, S},\Theta, \Sigma, Z^S, \boldsymbol{\pi}\}$ given the integration of node variables and network data is: $P(\mathbf{X},\mathbf{B}|\mathcal{M})
    =  P(\mathbf{X}|\mathcal{ A, S},\Theta, \Sigma, Z^S) P(\mathbf{B}|\mathcal{ A, S},\boldsymbol{\pi})$.
%
%\vspace{-0.05in}
%\begin{equation}
%    P(\mathbf{X},\mathbf{B}|\mathcal{M})
%    =  P(\mathbf{X}|\mathcal{ A, S},\Theta, \Sigma, Z^S) P(\mathbf{B}|\mathcal{ A, S},\boldsymbol{\pi})
%\end{equation}
%
With priors for parameters $\mathcal{M}$ the posterior likelihood is:
%\begin{equation}
 $   P(\mathcal{M}|\mathbf{X},\mathbf{B}) \propto P(\mathcal{M})P(\mathbf{X},\mathbf{B}|\mathcal{M})$.
%\end{equation}
%
  %
\section{Theory: Model Identifiability} \label{theory}
%\vspace{-0.05in}
Our method uses network data to avoid extra structural assumptions. In this section we formalize this idea through the identifiability of the proposed model. This property is important for interpretability of learned modules. Module networks and generally multivariate normal models for object variables can be un-identifiable, and imposing extra structural assumptions is necessary to overcome this. Here, we illustrate that the integrated learning proposed in this paper resolves the un-identifiability issue. First, we show that modeling node variables alone is identifiable only under very specific conditions. Then, we will restate some results from \cite{latouche2011overlapping} on the identifiability of overlapping block models. Using this result we show the identifiability of the model under some reasonable conditions.
\begin{lem}\label{lem1}
{\bf Node Variables Model:} For the model of node-specific variables $\mathbf{X}$, if we have:\\
%
%\begin{align}
$P(\mathbf{X}|\{\mathcal{ A, S}\}',\Theta', \Sigma')=P(\mathbf{X}|\{\mathcal{ A, S}\},\Theta, \Sigma)$
%\end{align}
%
\begin{enumerate}
\item Then, we can conclude: $\mu'=\mu \ and \ \Sigma'=\Sigma$.
%\vspace{-2mm}
\item If we further assume $\{\mathcal{ A, S}\}=\{\mathcal{ A, S}\}'$ and that each module has at least two non parent nodes and $\sum_k |Pa_k| < N$ and the covariance matrix $\Sigma$ is invertible, we can conclude: $\Theta=\Theta'$, $W=W'$ (Proof in Appendix A).
\end{enumerate}
\end{lem}
%
%\begin{sketch}
%\begin{enumerate}
%\item Considering that distributions of $\mathbf{X}$ are multivariate Normal under both parameter sets, it is straight forward that the mean and covariance parameters of two Normals should be the same. This can be formally shown by finding maximum of the distribution and curvature at any point for both sides, hence, $\mu'=\mu$ and $\Sigma'=\Sigma$.
%\item From the identifiability of $\mu$ and $\Sigma$, it is sufficient to show that $\mu$ and $\Sigma$ uniquely define $\Theta$, $W$ given $\{\mathcal{ A, S}\}$. Starting from $\Gamma_c$, we can consider the following set of linear equations:
%    %
%    \begin{align*}
%    {\boldsymbol \mu_c}=  {\Gamma}_c {\boldsymbol \mu_{c}^R}
%    \end{align*}
%    %
%    This is a set of equations with $N$ equations and $\sum_k |Pa_k|$ unknowns. Hence, when $\sum_k |Pa_k| < N$ this set of linear equations will lead to a unique solution if a solution exists.
%
%    For the $\Sigma$, given that it is invertible, we have:
%    %
%    \begin{align}
%    \Sigma^{-1}=(I-W)^{T}(I-W)
%    \end{align}
%    %
%    Considering that parents have the same value for $W_{nr}$ for $\forall n \in M_k$. Then, we can simply find $W_{nr}$ by solving $|Pa_k|*{W_{nr}}^2=\Sigma^{-1}_{ij}$ where $i,j$ are two genes that are non parents and belong to the module $M_k$.
%\end{enumerate}
%%
%\end{sketch}
%

The above lemma provides identifiability for the case where the structure $\{\mathcal{ A, S}\}$ is assumed to be known. However, in the case where we don't have the structure, the parameterizations of multivariate normal ($\mu$ and $\Sigma$) can be written in multiple ways in terms of $\Theta$ and $\{\mathcal{ A, S}\}$. This is due to existence of multiple decompositions for the covariance matrix.
In the following, we will use a theorem for identifiability of overlapping block models from~\cite{latouche2011overlapping} which is an extension of the results in~\cite{allman2009identifiability}. The results provide conditions for overlapping stochastic block models to be identifiable.
\begin{thm}\label{thm1}
{\bf Network Data Model:} If we have $P(\mathbf{B}| \{{\mathcal A,S}\},\boldsymbol{\pi})=P(\mathbf{B}| \{{\mathcal A,S}\}',\boldsymbol{\pi}')$,
%
%\begin{align}
%P(\mathbf{B}| \{{\mathcal A,S}\},\boldsymbol{\pi})&=P(\mathbf{B}| \{{\mathcal A,S}\}',\boldsymbol{\pi}')
%\end{align}
%
then: $\{{\mathcal A,S}\}=\{{\mathcal A,S}\}'$ with a permutation and $\boldsymbol{\pi}=\boldsymbol{\pi}'$ (except in a set of parameters which have a null Lebesgue measure) (Proof in Appendix B).
\end{thm}
%
%\begin{sketch}
%Our relational data model is an overlapping stochastic block model, where the blocks are parents and modules, with a specific parametrization among the modules and parents. Hence, we have the identifiability using the Theorem 4.1 in ~\cite{latouche2011overlapping}.
%\end{sketch}
%

Using the above Theorem and Lemma \ref{lem1} we can have the following Theorem for the identifiability of the model.
\begin{thm}\label{thm2}
{\bf Identifiability of Model:} If we have:
%
%\begin{align}
$P(\mathbf{B}| \{{\mathcal A,S}\},\boldsymbol{\pi})=P(\mathbf{B}| \{{\mathcal A,S}\}',\boldsymbol{\pi}')$ and
$P(\mathbf{X}|\{\mathcal{ A, S}\}',\Theta', \Sigma')=P(\mathbf{X}|\{\mathcal{ A, S}\},\Theta, \Sigma)$
%\end{align}
%
with assuming that each module has at least two non-parent nodes and $\sum_k |Pa_k| < N$ and the covariance matrix $\Sigma$ is invertible, then: $\{{\mathcal A,S}\}=\{{\mathcal A,S}\}'$ with a permutation, $\boldsymbol{\pi}=\boldsymbol{\pi}'$ , $\Theta=\Theta'$ and $W=W'$ (except in a set of parameters which have a null Lebesgue measure) (Proof in Appendix C).
\end{thm}
%
%
%\begin{sketch}
%This theorem is an immediate result from combination of Theorem \ref{thm1} and Lemma \ref{lem1}.
%Using (\ref{relation}), according to Theorem \ref{thm1} we have: $\{{\mathcal A,S}\}=\{{\mathcal A,S}\}'$ with a permutation and $\boldsymbol{\pi}=\boldsymbol{\pi}'$. Now, knowing $\{{\mathcal A,S}\}=\{{\mathcal A,S}\}'$ and equation (\ref{variable}) we can apply Lemma \ref{lem1} leading to $\Theta=\Theta'$ and $W=W'$. This concludes the proof.
%\end{sketch}
%
%
This Theorem, states the theoretical effect of integrated modeling on identifiability of modular structures, given that the sum of number of parents is less than the number of nodes (as is common in gene regulatory networks).
%
%\vspace{-3mm}
\section{Parameter Estimation using RJMCMC} \label{sec_inference}
We use a Gibbs sampler to obtain the posterior distribution $P(\mathcal{M}|\mathbf{X},\mathbf{B})$
%
%\begin{equation}
%    P(\mathcal{ A, S},\Theta, \Sigma, \boldsymbol{\pi}|\mathbf{X},\mathbf{B})
%\end{equation}
%
and design Metropolis-Hastings samplers for each of the parameters $\Theta, \Sigma, \boldsymbol{\pi}$ conditioned on the other parameters and data $\mathbf{X},\mathbf{B}$.
%The Gibbs steps for these parameters are described in the next subsection.
We use Reversible-Jump MCMC \citep{green1995reversible} for sampling from conditional distributions of the assignment and structure parameters $\mathcal{ A, S}$.
%
%
% mu_R
%\vspace{-4mm}
\subsection{Learning Parameters $\Theta,\Sigma,Z^S,\boldsymbol{\pi}$.}
%
%\vspace{-0.05in}
%Updating condition-specific parameters, $\theta_c \in \Theta$, for $c=\{1,...,C\}$, consists of updating variable means $\boldsymbol{\mu}_c$ and mixture matrix $\Gamma_c$.
%
To update the means, we only need to sample one value for means of parents assigned to the same module. This set of means of distinct parents $\boldsymbol{\mu}_c^{\bf{R}}$ are sampled with a normal proposal (Algorithm \ref{alg1}). %$Q_{\mu}(\boldsymbol{\mu}_c^{{\bf{R}}^{(i+1)}}|\boldsymbol{\mu}_c^{{\bf{R}}^{(i)}}) \sim \mathcal{N}(\boldsymbol{\mu}_c^{{\bf{R}}^{(i)}},I)$. The means of all expressions $\boldsymbol{\mu}_c^{(i+1)}$ and $\Theta^{(i+1)}$ are then computed accordingly.
%%%%%%%%%
%
% Gamma
Similarly we sample the parameters ${\boldsymbol\gamma_{c}^r}$, $z_k^{r}$ and $\pi_k^r$, corresponding to parent $r \in Pa_k$ of module $k$, from normal distributions. The conditions required for identifiability (from Theorem \ref{thm1}) are enforced in each iteration, such that samples violating the conditions are rejected. To update covariance $\Sigma$, each distinct element of the regression matrix $W$ corresponding to a module $k$, denoted as $w_k$, is updated.
%
%\vspace{4mm}
%\end{wrapfigure}
%
%For updating mixture coefficients $\Gamma_c$, we sample each of its distinct elements ${\boldsymbol \gamma_{c}^r}$, which corresponds to parent $r \in Pa_k$ of module $k$, also from a normal proposal. %$Q_{\Gamma}(\gamma_{c}^{r^{(i+1)}}|\gamma_{c}^{r^{(i)}})=\mathcal{N}(\gamma_{c}^{r^{(i)}},1)$.
%%The matrix $\Gamma_c^{(i+1)}$ is then updated based on assignment $\mathcal{A}^{(i)}$, dependencies ${S}^{(i)}$ and split-points $Z^{S}$ which define the contexts. Then, variable means $\boldsymbol{\mu}_c^{(i+1)}$ and $\Theta^{(i+1)}$ are updated from the mixture coefficients.
%%
%Similarly, split-points $z_k^{r} \in Z^S$ for each parent $r$ are sampled from normal distributions.
%%, from which $\Theta$ is updated accordingly.
%%
%% W
%
%$Q_W (w_k^{r^{(i+1)}} | w_k^{r^{(i)}}) \sim \mathcal{N}(w_k^{r^{(i)}},1)$.
%%The proposal covariance $\Sigma^{(i+1)}$ is then computed based on $W^{(i+1)}$ using equation (\ref{sigma}).
%% pi
%Sampling interaction parameters $\pi_k^r$ is also performed using a normal distribution proposal.
%$Q_{\pi}(\pi_k^{{r}^{(i+1)}}|\pi_k^{r^{(i)}})=\mathcal{N}(\pi_k^{{r}^{(i)}},1)$.
Due to the symmetric proposal distribution, the proposal is accepted with probability $P_{mh} = \min \{1,{P(\mathcal{M}^{(i+1)}|X,B)\over P(\mathcal{M}^{(i)}|X,B)}\}$
%
%\vspace{-0.05in}
%\begin{equation}\label{mh}
%  P_{mh} = \min \{1,{P(\mathcal{M}^{(i+1)}|X,B)\over P(\mathcal{M}^{(i)}|X,B)}\}
%\end{equation}
%\vspace{-0.05in}
%
where $\mathcal{M}^{(i)}=\{ \mathcal{A,S},\Theta,\Sigma,Z^S \boldsymbol{\pi}\}^{(i)}$.
%Gibbs iterations are repeated until the samples are mixed properly.
%
%
\begin{algorithm}[h]
   \caption{RJMCMC for sampling parameters}
   \label{alg1}
\begin{algorithmic}
   \STATE {\bfseries Inputs:} \\
   Node Variables Data $\mathbf{X}$\\
   Network Data $\mathbf{B}$\\
  % \STATE Initialize parameters
   %\REPEAT
   \FOR{iterations $i=1$ {\bfseries to} $I$}
   \STATE Sample $\mathcal{A}^{(i+1)}$ given $\mathcal{A}^{(i)}$ using Alg 2 in appendix
   \STATE Sample $\mathcal{S}^{(i+1)}$ given $\mathcal{S}^{(i)}$ using Alg 3 in appendix
   \FOR{modules $k=1$ {\bfseries to} $K^{(i)}$}
   \STATE Propose $w_k^{(i+1)}  \sim \mathcal{N}(w_k^{(i)},I)$
   \STATE Accept with probability $P_{mh}$; update $\Sigma^{(i+1)}$
   \FOR{parents $r=1$ {\bfseries to} $R_k$}
    \STATE Propose $z_k^{r (i+1)} \sim \mathcal{N}(z_k^{r(i)},I)$; accept with $P_{mh}$
    \STATE Propose $\pi_k^{r (i+1)}\sim \mathcal{N}(\pi_k^{r (i)},I)$; accept with $P_{mh}$
    \ENDFOR
   \ENDFOR
   \FOR{condition $c=1$  {\bfseries to} $C$}
   \STATE Propose $\boldsymbol{\mu}_c^{{\bf R}(i+1)} \sim \mathcal{N}(\boldsymbol{\mu}_c^{{\bf R}(i)},I)$; accept with $P_{mh}$
   \STATE Propose ${\boldsymbol\gamma_{c}^{{\bf R}(i+1)}} \sim \mathcal{N}( {\boldsymbol\gamma_{c}^{{\bf R}(i)}} ,I)$; accept with $P_{mh}$
   \ENDFOR %conditions
   %
%   \FOR{module $k=1$  {\bfseries to} $K$}
%   \FOR{parent $r=1$  {\bfseries to} $R_k$}
%
%   \ENDFOR % parents
%   \ENDFOR % module
   %\IF{}
   %\ENDIF
   \ENDFOR % Gibbs
   %\UNTIL{$noChange$ is $true$}
\end{algorithmic}
\end{algorithm}
%
% \begin{minipage}[t]{0.45\textwidth}
%    \vspace{-6mm}
%    \flushleft
%\begin{algorithm}[H]
%   \caption{RJMCMC to update parameters} % from ${\mathcal{A}}^{(i)}$}
%   \label{alg2}
%\begin{algorithmic}[1]
%%\STATE {\bfseries Inputs:} $\mathbf{X},\mathbf{B}$; $\mathcal{A}^{(i)},\mathcal{S}^{(i)},\Theta^{(i)}, \Sigma^{(i)}, \boldsymbol{\pi}^{(i)}$
%   %\STATE Set $p_{-1},p_{0},p_{+1}$
%   \STATE $\Theta,\Sigma,Z^S,\boldsymbol{\pi}$
%   \FOR{sample $i=1$  {\bfseries to} number of iterations }
%   \STATE Propose $\boldsymbol{\mu}_c^{{\bf{R}}^{(i+1)}} \sim \mathcal{N}(\boldsymbol{\mu}_c^{{\bf{R}}^{(i)}},I)$
%   \STATE Accept with $P_{mh}$
%
%    \STATE Propose $\pi_k^r$
%   \ENDFOR
%\end{algorithmic}
%\end{algorithm}
%\vspace{-3mm}
%\end{minipage}
%
%\vspace{-1mm}
%\subsection{Learning Module Assignment $\mathcal{A}$.}
\subsection{Learning assignments $\mathcal{A,S}$.}
Learning the assignment of each node to a module, involves learning the number of modules. Changing the number of modules however, changes dimensions of the parameter space and therefore, densities will not be comparable. Thus, to sample from $P(\mathcal{ A}|\mathcal{S},\Theta, \Sigma,,Z^S \boldsymbol{\pi},\mathbf{X},\mathbf{B})$, we use the Reversible-Jump MCMC method \citep{green1995reversible}, an extension of the Metropolis-Hastings algorithm that allows moves between models with different dimensionality.
%$$\vspace{-0.8in}$$
%
%\vspace{-3mm}
%\vspace{-12mm}
%\end{wrapfigure}
%
%
%
In each proposal, we consider three close move schemes
%on assignment function $\mathcal{A}$
of increasing or decreasing the number of modules by one, or not changing the total number. For increasing the number of modules, a random node is moved to a new module of its own and for decreasing the number, two modules are merged. In the third case, a node is randomly moved from one module to another module, to sample its assignment (Algorithm \ref{alg2} in Appendix D).
To sample from the dependency structure (assignment of parents) $P(\mathcal{ S}|\mathcal{A},\Theta, \Sigma,Z^S \boldsymbol{\pi},\mathbf{X},\mathbf{B})$, we also implement a Reversible-Jump method, as the number of parents for each module needs to be determined. Two proposal moves are considered for $\mathcal{S}$ which include increasing or decreasing the number of parents for each module, by one (Algorithm \ref{alg3} in Appendix E).
\section{Results}
\subsection{Synthetic Data}
We first tested our method on synthetic node-variables and network data generated from the proposed model. A dataset was generated for $N=200$ nodes in $K=4$ modules with $C=50$ conditions for each node variable. Parents were assigned from a total of $R=10$ number of candidates. Parameters $\pi$, $\gamma$ and $W$ were chosen randomly, preserving parameter sharing of modules. The inference procedure was run for 20,000 samples. Exponential prior distributions were used for number of parents assigned to each module, to avoid over-fitting.
\begin{figure}[h]
%\vspace{-4mm}
\center
  %\flushleft
  % Requires \usepackage{graphicx}
  \includegraphics[width=1\textwidth]{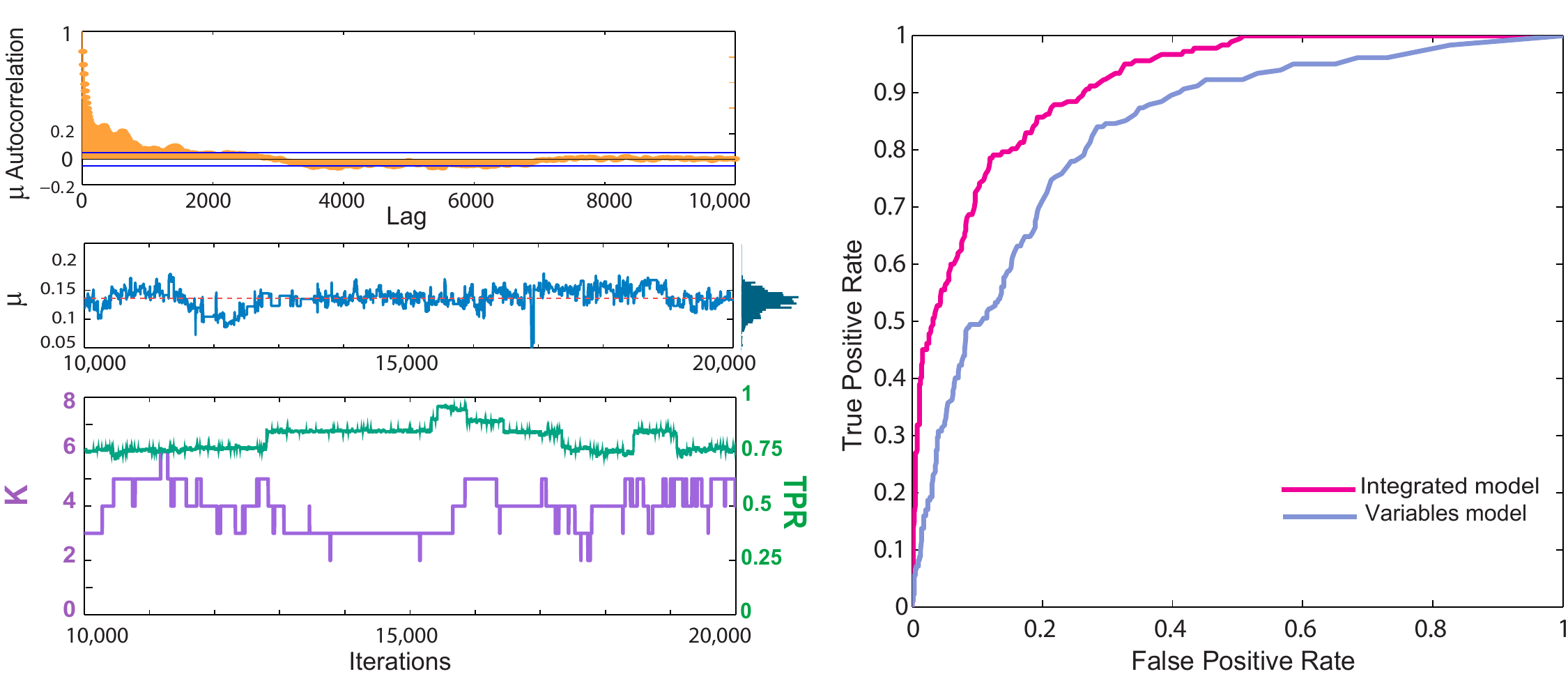}
  %\vspace{-8mm}
 % \vspace{-3mm}
  \caption{{\bf Results for sythetic data:}  Autocorrelation for an example variable mean (top); gibbs samples and posterior after burn-in period (actual mean shown with red line); number  of modules (purple) and true positive rate of recovered links (green), ROC curve for integrated model and variables model (bottom)}
  \label{fig_syn1}
  %\vspace{-5mm}
  \end{figure}
Figure \ref{fig_syn1} shows the autocorrelation for samples of variable mean $\mu_c^n$ for an example gene. The samples become independent after a lag and thus we removed the first $10,000$ iterations as burn-in period. Samples from posteriors, including the number of modules $K$, exhibit standard MCMC movements around the actual value (actual $K=4$). We also calculated the true positive rate and false positive rates based on actual dependency links.
We repeated the estimation of true positive and false positive rates for 100 random datasets with the same size as mentioned and computed the average ROC for the model (figure \ref{fig_syn1}). As comparison, for each generated dataset, we also tested the sub-model for variable data (excluding the model for network data) to infer links (figure \ref{fig_syn1}). We performed bootstrapping on sub-samples with size $1000$ to compute variance of AUC (area under curve) and paired t-tests confirmed improved performance of integrated model compared to the variables sub-model  ($p<0.05$).

The parameter sharing property in modular structures allows parallel sampling of parameters $w_k$ and $\gamma_{(k)}^r$, $z_k^r$,$\pi_k^r$ for each module $k$, in each iteration and in different conditions. We used Matlab-MPI for this implementation. It takes an average of $36 \pm 8$ seconds to generate 100 samples for $N=200$, $C=50$, $R=10$ on an i5 $3.30$GHz Intel(R). For further enhancement, module assignments were initialized by k-means clustering of variables.

\subsection{M. tuberculosis Gene Regulatory Network}

\begin{figure}[h]%[width=0.5\textwidth]
%\begin{figure}
  \center
  %\vspace{-5mm}
  % Requires \usepackage{graphicx}
  \includegraphics[width=0.9\textwidth]{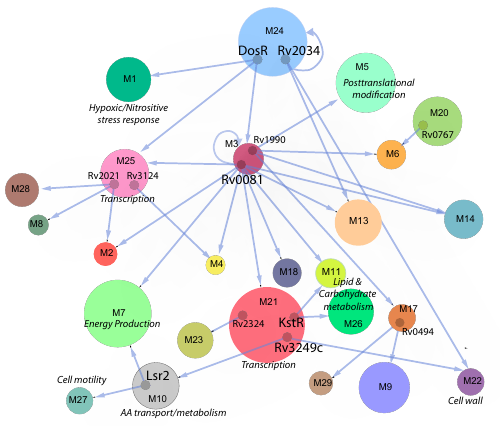}\\
  \caption{{\bf Regulatory structures between largest modules inferred for MTB}: Regulators assigned to each module are shown; the size of circles are proportional to number of genes assigned to the module. Enriched functional annotations are highlighted (details in table \ref{tab1}).}
  \label{fig_tbnet}
  %\vspace{-5mm}
%\end{wrapfigure}
\end{figure}

%Inferred modules for MTB are enriched for biological functions
We applied our method to identify modular structures in the Mycobacterium tuberculosis (MTB) regulatory network. MTB is the causative agent of tuberculosis disease in humans and the mechanisms underlying its ability to persist inside the host are only partially known \citep{flynn2001tb}.
%Experiments on MTB are time-consuming, difficult and require special facilities and computational models for regulation of MTB genes can help our understanding of TB-host interactions and pathogenicity and ultimately guide experiments towards development of new drugs and vaccines.
We used interaction data identified with ChIP-Seq of 50 MTB transcription factors and expression data for different induction levels of the same factors in 87 experiments, from a recent study by \cite{Galagan2013}. Only bindings of factors to upstream intergenic regions were considered. We tested our method on $3072$ MTB genes which had binding from at least one of these factors
%
%The microarray expression dataset contained 87 experiments, includes expression measurements when a TF is induced and over-expressed with different induction levels.
%
and performed 100,000 number of iterations on the combination of the two datasets.
%
%Out of the total genes, 815 could be assigned as a member of a module with high confidence (posterior probability of assignment$>0.9$). Genes with low assignment probability might be regulated by any of the other $\sim$150 MTB factors for which ChIP-Seq data is not available yet.
For each gene, we inferred the mode of its assignments to modules (after removing burn-in samples) and obtained 29 modules in total. The largest modules and the assigned regulators are shown in figure \ref{fig_tbnet}.
%The majority of idenitifed modules, have only one regulator assigned, as expected from the binding network \citep{Galagan2013}.
%
%
%
%\begin{figure}[h]%[width=0.5\textwidth]
%%\begin{figure}
%%\vspace{-5mm}
%  \flushleft
%  \centering
%  % Requires \usepackage{graphicx}
%  \includegraphics[width=0.5\textwidth]{dosR3.pdf}
%  %\vspace{-5mm}
%  \caption{Inferred regulatory program for module M25 of fig. \ref{fig_tbnet} showing the induction of these genes by DosR is mediated through Rv0081 in context (c)}
%  \label{fig_dosr}
% % \vspace{-9mm}
%%\end{wrapfigure}
%\end{figure}

We found functional enrichment of modules using Gene Ontology (GO) terms and COG category annotations from the TBDB database \citep{reddy2009tb} (enrichments indicate higher probability of observing a function in module compared to other modules). Out of 29 modules, 26 were enriched for at least one COG category with Bonferroni corrected $p<0.05$. The enrichments for the top major identified modules are shown in table \ref{tab1}.
For each module, the number of assigned genes and examples of previously studied genes are presented. The identified regulators of each module and enriched annotations confirm known functions for some regulators, such as the role of KstR (Rv3574) in regulating lipid metabolism \citep{kendall2007highly}, confirmed in modules M26 and M11; and DosR (Rv3133c) in nitrosative stress response \citep{voskuil2003inhibition} (module M1) and transcription \citep{rustad2008enduring} (module M25). Novel functions for other regulators and the combinations of regulators acting together are also presented.

\begin{table*}[t]
\scalebox{0.75}{
\begin{tabular}{p{1cm}p{1.2cm}p{3.7cm}p{2.5cm}p{5cm}p{4.3cm}}
  \toprule
  % after \\: \hline or \cline{col1-col2} \cline{col3-col4} ...
  Module ID & Number of Genes & Example Genes Assigned to Module &  Regulators & Enriched COG Catergories ($p<0.05$)& Enriched GO terms ($p<0.05$) \\
  \midrule \midrule
   M21& 291 & KstR, Rv3249c, sigI, relA, helZ, recG & Rv0081 & Replication, recombination and repair;
Transcription  &  extracellular region; growth; plasma membrane\\
   \midrule
   M24 & 258 & DosR, sigA, sigL, clpP1,2 & Rv2034 &Intracellular trafficking, secretion, and vesicular transport;
Secondary metabolites biosynthesis, transport and catabolism & extracellular region; plasma membrane  \\
   \midrule
  M7 & 250 & Rv0324, sigE, rpoA, icl, sucC, narK1, nuoAB, nuoDEFG & Rv0081, Lsr2 & Energy production and conversion;
Inorganic ion transport and metabolism & NADH dehydrogenase (ubiquinone) activity; growth; plasma membrane  \\
  \midrule
   M5 & 214 & inhA, fabH & Rv1990c &Posttranslational modification, protein turnover, chaperones&  growth; plasma membrane \\
  \midrule
   M25& 161 & ideR, sigB, nusG, argR, lipP, Rv2021c, Rv3124  &Rv0081, DosR & Transcription; Defense mechanisms & plasma membrane;
succinate dehydrogenase activity   \\
  \midrule
   M10 & 154 & lysA, dapF, fprA, lipO, fadD7, fadD30, fadA6 & Rv3249  & Amino acid transport and metabolism; &  plasma membrane \\
   \midrule
   M26 & 148 & sugA,B,C; mutA,B & KstR & Carbohydrate transport and metabolism;
Lipid transport and metabolism & growth; propionate metabolic process, methylmalonyl pathway \\
   \midrule
   M1 & 144 & fabG4, fadD8 & DosR &Secondary metabolites biosynthesis, transport and catabolism & cellular response to nitrosative stress;
growth; plasma membrane\\
      \midrule
   M22 & 60 & fas, fadA4, pcaA, metB  &  Rv3249c, Rv2034 & Cell wall/membrane/envelope biogenesis & plasma membrane \\
   \midrule

   M27 & 59 & kasA-B, fabD, accD6  & Lsr2  & Cell motility &  plasma membrane \\
   \midrule
   M11 & 48 & fadA3, fadD4, lipC, lipW, nuoH-N, narI,J,H & Rv0081, KstR  & Energy production and coversion; Lipid transport and metabolism &
   NADH dehydrogenase (ubiquinone) activity; nitrate reductase activity\\ %  (narI,J,G,H) (mmaA1,2,3,4)
      \midrule
   M3 & 36 & Rv0081, Rv0232, Rv1990c, fadE4, fadE5 & DosR & Energy production and conversion & - \\
  % &  &  &  \\
  \bottomrule
\end{tabular}}
\caption{Enrichment of functional annotations for largest modules controlled by major MTB regulators}
\label{tab1}
%\vspace{-2mm}
\end{table*}

\begin{figure}[t]%[width=0.5\textwidth]
%\begin{figure}
%\vspace{-5mm}
 % \flushleft
  \centering
  % Requires \usepackage{graphicx}
  \includegraphics[width=0.5\textwidth]{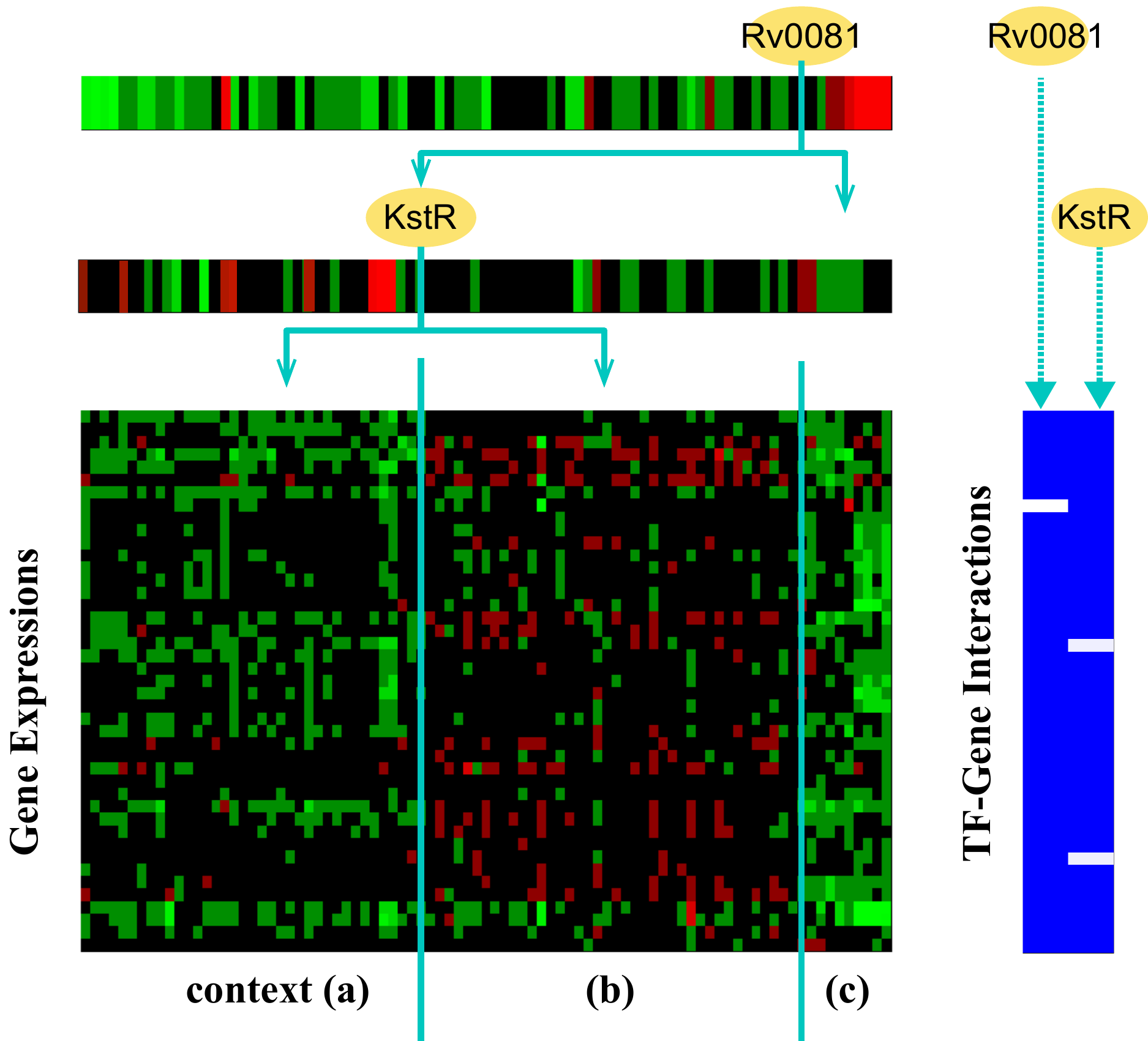}
  %\flushright
  \includegraphics[width=0.48\textwidth]{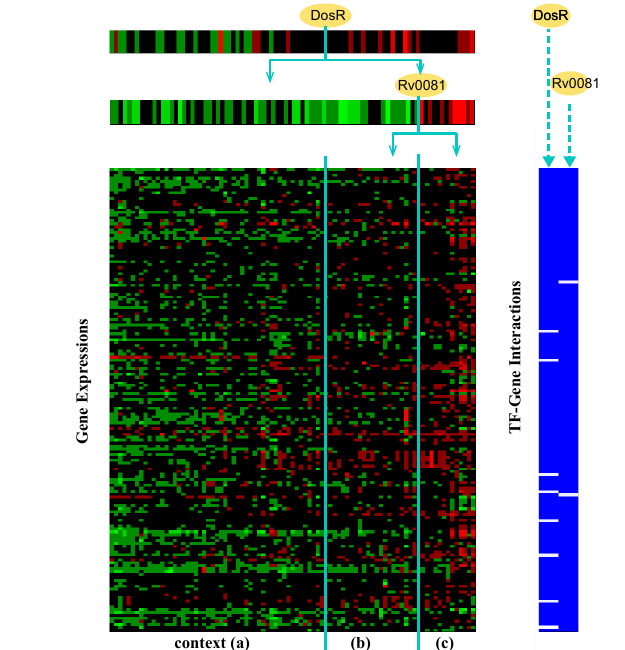}
  %\vspace{-3mm}
  \caption{Examples of inferred regulatory programs: (Left) module M11 of fig. \ref{fig_tbnet} showing that either of Rv0081 and KstR can repress the module in contexts (a) and (c); (Right) module M25 of fig. \ref{fig_tbnet} showing the induction of these genes by DosR is mediated through Rv0081 in context (c)}
  \label{fig_kstr}
  %\vspace{-2mm}
%\end{wrapfigure}
\end{figure}

As shown in figure \ref{fig_tbnet}, many modules are controlled by more than one regulator, highlighting the significance of combinatorial regulations in controlling gene expressions in this network.
 %For modules with a single regulator, activation or repression signs were inferred based on estimated coefficients $\gamma^r$.
The inferred structure identifies multiple feed-forwards loops (FFLs), many of which involve a hub regulator Rv0081 and another regulator.
 %For example M25 is regulated by two hypoxic regulators DosR and Rv0081, M13 is regulated by Rv0081 and Rv2034, etc. The regulator Rv0081 has not been well-studied and was recently identified as a major hub, binding to more than a 1000 genes in the MTB genome \citep{Galagan2013}.
 FFLs are known to lead to dynamic transient responses or time delays in gene expression \citep{mangan2003structure} and the role of Rv0081 in driving multiple FFLs in MTB can be further studied.
% In addition to the mentioned feed-forward loop, there a feedback link is inferred from Rv3249c (predicted as a repressor) back to the module containing DosR.
Also, two auto-regulating feedbacks were inferred from Rv0081 to its module M3, and from Rv2034 to M24, which may contribute to stabilizing and noise-reduction \citep{kaern2005stochasticity} in transcription of the hub regulators.
One inferred module is M11 shown in figure \ref{fig_kstr} which is regulated by Rv0081 and KstR (Rv3574). KstR is known to be involved in cholesterol and lipid catabolism \citep{kendall2007highly} and the module is enriched for "Energy production and conversion" and "Lipid transport
and metabolism" COG categories (table \ref{tab1}). The inferred program depicted in figure \ref{fig_kstr} shows that either of the two regulators can repress the expression of the 48 genes assigned to this module, which include lipases and genes involved in fatty acid $\beta$-oxidation and triacylglycerides cycle metabolic pathways. KstR itself is also regulated by Rv0081, forming another FFL and the roles of both factors in repressing these pathways can be further investigated. Thus, a hypoxic (oxygen deprivation) regulator Rv0081, regulates lipid metabolism genes through KstR. The two factors of hypoxic adaptation and lipid catabolism are two main factors involved in MTB persistence \citep{flynn2001tb,Galagan2013}.

Figure \ref{fig_kstr} shows module M25 containing 161 genes, with an interesting regulatory program involving two MTB hypoxic adaptation regulators: Rv3133c (DosR) and Rv0081. DosR is well known to activate the initial response of MTB in hypoxic conditions \citep{park2003dosr}. As table \ref{tab1} shows, M25 is enriched for "Transcription" in COG categories. The genes assigned to this module include other regulators such as Rv2021c, Rv3124 known to be induced in later time points (after 24 hours) in hypoxia. The mechanism driving this enduring hypoxic response is not well known  \citep{rustad2008enduring}.
The inferred regulatory program for this module predicts induction of most genes in the module in conditions where both DosR and Rv0081 are expressed (context (c) in figure \ref{fig_kstr}). This combinatorial regulation could be acting as either a logical AND gate, where both factors are required, or Rv0081 might be the only necessary activator of the module. However, Rv0081 itself is also regulated by DosR, which creates a feed-forward loop structure driving this module (see figure \ref{fig_tbnet}).
%The TF-gene interaction profile (heatmap on right) shows high probability of binding of both regulators to this module.
Hence, this program illustrates the significance of Rv0081 and DosR in the form of a FFL in mediating the induction of a second hierarchy of regulators with a time delay, leading to a later hypoxic response.
%
%\vspace{-4mm}
%\subsection{Integrative learning reduces false positive regulatory links}
%

We showed in section \ref{theory} that integration of network data has theoretical advantages in terms of model identifiability. Here, we show that it can also reduce the number of false positive regulatory links in MTB data.
As a gold standard, we used previously validated links (by EMSA, RTq-PCR) for two MTB regulators, including 48 known links for DosR from \cite{voskuil2003inhibition} and 72 known links for KstR from \cite{kendall2007highly}.
 %The regulons were analyses, microarray studies, and comparative genomics.
%
We calculated the area under precision-recall for our method by comparing posterior probabilities for DosR and KstR links to known links (table \ref{tab2}). As comparison, we also applied common methods shown to have best performance in DREAM challenge contests \citep{marbach2012wisdom} in inferring regulatory networks from gene expression only. These include Mutual Information between expression profiles (MI), CLR \citep{faith2007clr}and GENIE3 \citep{irrthum2010genie}.
%, and Network Deconvolution (ND) for filtering indirect links \citep{feizi2013network}.
We applied these on the above MTB expression data, and compared the inferred links to the gold standard set. As the number of validated links in MTB are small, we also scored the predictions from co-expression methods to the MTB ChIP-Seq data \citep{Galagan2013} for the same two regulators. Also, none of these methods assume modular structures.
%We did not compare our inferred links to ChIP-Seq as it was part of the training data.

\begin{table}
%\vspace{-3mm}
\caption{Area under precision-recall AUPR($\%$) calculated for link prediction using proposed method and other common co-expression methods, applied to MTB data. The predictions are scored vs known and ChIP-Seq links for two regulators }
%\vspace{1mm}
\center
\scalebox{0.9}{
\begin{tabular}{p{4cm}p{1.1cm}p{1.1cm}p{1.1cm}p{1.1cm}}
  \toprule
  % after \\: \hline or \cline{col1-col2} \cline{col3-col4} ...
  Gold Standard  & \multicolumn{2}{c}{Validated Links}  & \multicolumn{2}{c}{ChIP-Seq Links} \\
Regulator  & DosR  & KstR  &  DosR &  KstR
  \\ No. of Targets & (48) & (72) & (528) & (503)\\
%  \\Method & &  & &    \\
%  Method&
  \midrule
  % \midrule
%  EGRIN? \\
%  \midrule
%  Module Network\\
\midrule
MI & 39.04 & 9.24 & 25.00 & 17.85\\
  \midrule
  CLR & 48.25 & 9.37 & 21.44 & 16.77\\
 %  \midrule
%  CLR + ND & 36.55 & 6.07\\

   \midrule
  GENIE3 & {\bf 62.26} & 31.37 & 21.55 & 19.44\\
%   \midrule
%  MI + ND & 40.04 & 9.64 & 28.95 & 17.16\\
%  \midrule
%  GENIE3 + ND & 32.41 & 24.86 & 17.08 & 18.20\\
  \midrule
  Proposed Model & {\bf 72.13} & {\bf 65.72} & {\bf 79.62} &{\bf  70.06}
  \\
  \bottomrule
\end{tabular}}
\label{tab2}
%\vspace{-5mm}
\end{table}
We then applied Module Networks \citep{Segal2005} to the same expression dataset and compared predictions to known links and ChIP-Seq data (table \ref{tab3}). We set the maxmimum number of modules to 10 and constrained the candidate pool of regulators to the 50 ChIPped regulators only.
%The method identified 2401 TF-gene interactions, out of which only 215 (precision: 8.9$\%$) had ChIP evidence for binding to upstream or genic regions of genes and only 18.53$\%$ of genes had binding from at least one of the regulators assigned to their module.
 On average $2.8 \pm 0.63$ number of regulators were assigned to each module, with a mode of 3, whereas the ChIP-Seq network shows a mode of 1 for in-degree of genes~\citep{Galagan2013}, i.e. most genes have only one regulator binding.
 As the predicted links from module networks are deterministic, an AUPR score can not be reported, thus we compared to precision and recall of posterior mode from our models.  Note small precision values are due to small number of validated links, i.e. if a link is not validated experimentally it may not be wrong.
For a fair comparison of models without the effect of interaction data, we also compared to performance of our model for variables data only (table \ref{tab3}).
%
%As a result, 4264 interactions were identified, out of which 739 (17.33\%) had binding evidence, and 32.76\% of genes had binding from at least one of the regulators assigned to their module.
%
These results show that module networks and in general co-expression methods have many false positives and integrating interaction data is necessary for inference of direct regulatory relationships.
\begin{table}[h]
%\vspace{-3mm}
%\vspace{-3mm}
\caption{Percentage of Precision (P) and Recall (R) for link prediction using module networks and proposed models.}
%\vspace{-1mm}
\center
%\flushleft
\scalebox{0.85}{
\begin{tabular}{p{6cm}p{1cm}p{1cm}p{1cm}p{1cm}p{1cm}p{1cm}p{1cm}p{0.8cm}}
  \toprule
  % after \\: \hline or \cline{col1-col2} \cline{col3-col4} ...
  Gold Standard  & \multicolumn{4}{c}{Validated Links}  & \multicolumn{4}{c}{ChIP-Seq Links} \\
Regulator  & \multicolumn{2}{c}{DosR}   & \multicolumn{2}{c}{KstR}  & \multicolumn{2}{c}{ DosR} & \multicolumn{2}{c}{ KstR} \\
 % \\ No. Targets & (48) & (72) & (528) & (503)\\
 \midrule
& P & R & P & R & P & R & P & R \\
%  \\Method & &  & &    \\
%  Method&
  \midrule
  % \midrule
%  EGRIN? \\
%  \midrule
%  Module Network\\
\midrule
 % \midrule
  Module Networks & 3.8 & 81.2 & 6.5 & 86.1 & 40.1 & 76.3  & 35.8 & 67.4\\%& 2401 & 215 &
  \midrule
  Proposed Model for Variables (mode) & 4.6&77.1& 7.2 &77.8 & 55.0 & 83.7 & 52.5& 80.5 \\%& 4264 & 739
  \midrule
  Proposed Integrated Model (mode)& 6.5&89.6 & 10.6&84.7 & 75.4&93.4&83.6&95.6\\
  \bottomrule
\end{tabular}}
%\vspace{1mm}
\label{tab3}
%\vspace{-2mm}
\end{table}

%\end{methods}
%
%
%\vspace{-3mm}
%
\subsection{Twitter Network}
\begin{figure}[h]
%\vspace{-4mm}
\center
%  \flushleft
  % Requires \usepackage{graphicx}
  \includegraphics[width=0.9\textwidth]{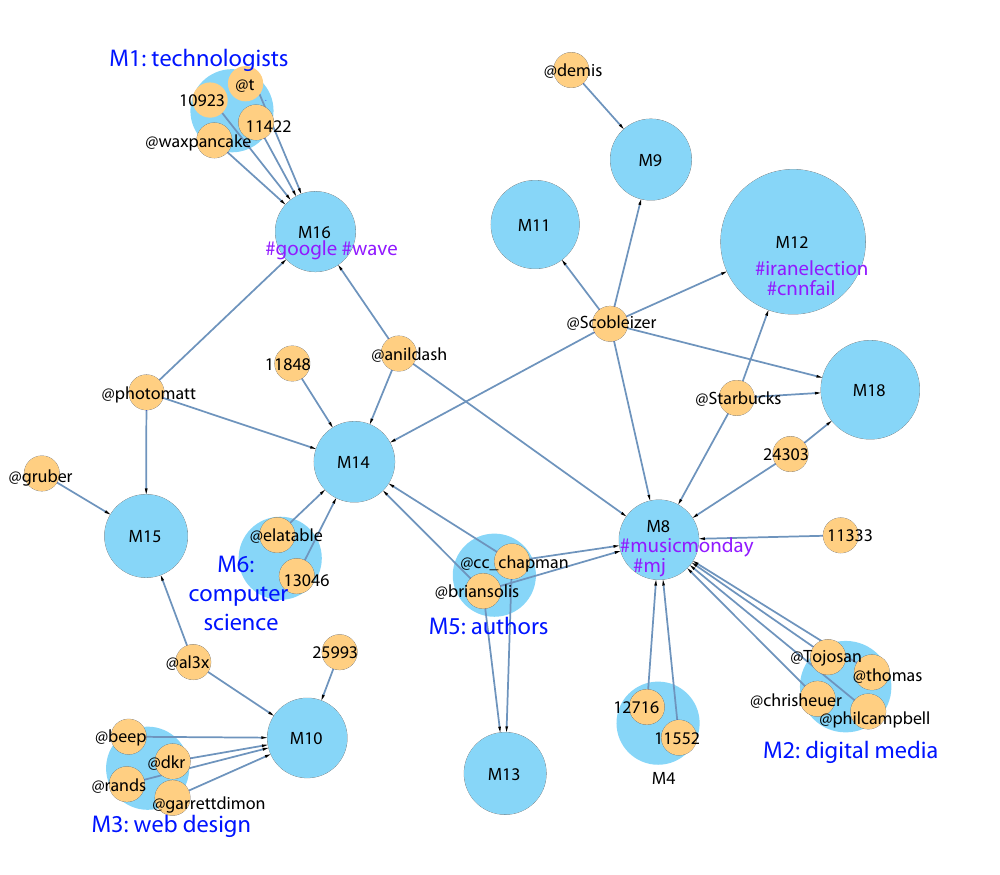}
  %\vspace{-8mm}
  %\vspace{-7mm}
  \caption{{\bf Influence structures inferred for a twitter subnetwork}: Top posted hashtags and user interests are highlighted}
  \label{fig_twitter}
  %\vspace{-7mm}
 \end{figure}

As a second application, we used our method to find influence structures in a social network.
In social networks such as twitter, the activity of users, e.g. number of tweets posted by a user in a time window, can be influenced by other users. To find these influence patterns, one approach would be to search for all other users that have correlated activity, e.g. same number of posts in a day. However, given that users are more likely to be influenced by users whom they are following, integrating the social graph of who-follows-who would improve accuracy and speed in finding influential users that affect a large community (module) of users. We applied our method on integration of two types of data from twitter. Number of user posts (tweets) are considered as node variables, time windows of one day are considered as conditions, and the network of followings is considered as network information. The dataset of tweets during a period of 4 months from (June to Sept 2009) \citep{yang2011twitter} was combined with the social graph of who-follows-who \citep{Kwak10twittergraph} and 450 number of users were randomly selected that had data in both datasets. Figure \ref{fig_twitter} shows the inferred modular structures of influence between users with circle sizes proportional to number of users assigned to each module.

The results present interesting structures of influence for each community. For example, module (community) M13 is influenced by best-selling authors such as Brian Solis and C.C. Chapman which are assigned to M5, while users in module M10 are mostly influenced by well-known web designers and developers in M3 including Ethan Marcotte ($@$beep), Garett Dimon and Michael Lopp ($@$ranks). Module M16 is mostly influenced by famous technologists including Tantek Celik ($@$t) and Andy Baio ($@$waxpancake).
Module M6 contains computer scientists such as Bradley Horowitz ($@$elatable).
The most influential users (largest fan-out degrees) in this subnetwork were famous blogger Robert Scoble ($@$Scobleizer) and $@$Starbucks. These results also identify communities with diverse interests such as M8, i.e. follow users belonging to diverse communities.
The top hashtags posted by users of each community is consistent with interests and professions of their influencers and highlights major events of that period in 2009, such as launch of Google Wave, Iran election and  Michael Jackson's death (figure \ref{fig_twitter}). Thus this method can identify communities with common influencers in social networks.
%The largest module (community) m12 contains 139 users which are influenced by user IDs , meaning these users follow user X and they are likely to post tweets around the same time as the influencer.
%\vspace{-5mm}
%
%
%\vspace{-5mm}
\section{Conclusion}
We proposed a model for learning dependency structures between modules, from network data and node variables.
We showed that the assumption of shared parents and parameters for nodes in a module, together with integration of network data deals with under-determination and un-identifiability, improves statistical robustness and avoids over-fitting.
%This integration improves accuracy and avoids over-fitting.
%
 We presented a reversible-jump inference procedure for learning model posterior. %
 Our results showed high performance on synthetic data and interpretable structures on real data from {\it M. tuberculosis} gene network and twitter social network. Results for MTB gene regulatory network revealed feed-forward loops and insights into condition-specific regulatory programs for lipid metabolism and hypoxic adaptation. Inferred modules in a twitter subnetwork identified influencing users for different communities.

\section*{Acknowledgments}
We  acknowledge funding from 
 the Hariri Institute for Computing and Computational Science \& Engineering, 
 the National Institute of Health under grants HHSN272200800059C and R01 GM096193, 
 the National Science Foundation under grant IIS-1149662,
 the Army Research Office under grant MURI W911NF-11-1-0036, and 
 from an Alfred P. Sloan Research Fellowship.

% In the unusual situation where you want a paper to appear in the
% references without citing it in the main text, use \nocite
%\nocite{langley00}
%\begin{spacing}{0.8}
\bibliographystyle{plainnat}
%\bibliography{elhamlib}

\begin{thebibliography}{42}
\providecommand{\natexlab}[1]{#1}
\providecommand{\url}[1]{\texttt{#1}}
\expandafter\ifx\csname urlstyle\endcsname\relax
  \providecommand{\doi}[1]{doi: #1}\else
  \providecommand{\doi}{doi: \begingroup \urlstyle{rm}\Url}\fi

\bibitem[Airoldi(2007)]{Airoldi2007}
E.M. Airoldi.
\newblock Getting started in probabilistic graphical models.
\newblock \emph{PLoS Computational Biology}, 3\penalty0 (12):\penalty0 e252,
  2007.

\bibitem[Airoldi et~al.(2008)Airoldi, Blei, Fienberg, and
  Xing]{airoldi2008mixed}
E.M. Airoldi, D.M. Blei, S.E. Fienberg, and E.P. Xing.
\newblock Mixed membership stochastic blockmodels.
\newblock \emph{The Journal of Machine Learning Research}, 9:\penalty0
  1981--2014, 2008.

\bibitem[Airoldi et~al.(2013{\natexlab{a}})Airoldi, Costa, and
  Chan]{Airoldi_Costa_Chan_2013}
E.M. Airoldi, T.B. Costa, and S.H. Chan.
\newblock Stochastic blockmodel approximation of a graphon: {Theory} and
  consistent estimation.
\newblock In \emph{Advances in Neural Information Processing Systems (NIPS)},
  volume~26, pages 692--700, 2013{\natexlab{a}}.

\bibitem[Airoldi et~al.(2013{\natexlab{b}})Airoldi, Wang, and
  Lin]{Airoldi:2013fk}
E.M. Airoldi, X.~Wang, and X.~Lin.
\newblock Multi-way blockmodels for analyzing coordinated high-dimensional
  responses.
\newblock \emph{Annals of Applied Statistics}, 7\penalty0 (4):\penalty0
  2431--2457, 2013{\natexlab{b}}.

\bibitem[Allman et~al.(2009)Allman, Matias, and
  Rhodes]{allman2009identifiability}
E.S. Allman, C.~Matias, and J.A. Rhodes.
\newblock Identifiability of parameters in latent structure models with many
  observed variables.
\newblock \emph{The Annals of Statistics}, 37\penalty0 (6A):\penalty0
  3099--3132, 2009.

\bibitem[Aral et~al.(2009)Aral, Muchnik, and
  Sundararajan]{aral2009distinguishing}
S.~Aral, L.~Muchnik, and A.~Sundararajan.
\newblock Distinguishing influence-based contagion from homophily-driven
  diffusion in dynamic networks.
\newblock \emph{Proceedings of the National Academy of Sciences}, 106\penalty0
  (51):\penalty0 21544--21549, 2009.

\bibitem[Azari~Soufiani and Airoldi(2012)]{azarigraphlet}
H.~Azari~Soufiani and E.M. Airoldi.
\newblock Graphlet decomposition of a weighted network.
\newblock \emph{Journal of Machine Learning Research}, \penalty0 (W\&CP 22
  (AISTATS)):\penalty0 54--63, 2012.

\bibitem[Celniker et~al.(2009)Celniker, Dillon, Gerstein, Gunsalus, Henikoff,
  Karpen, Kellis, Lai, Lieb, MacAlpine, et~al.]{celniker2009unlocking}
S.E. Celniker, L.~Dillon, M.B. Gerstein, K.C. Gunsalus, S.~Henikoff, G.H.
  Karpen, M.~Kellis, E.C. Lai, J.D. Lieb, D.M. MacAlpine, et~al.
\newblock Unlocking the secrets of the genome.
\newblock \emph{Nature}, 459\penalty0 (7249):\penalty0 927--930, 2009.

\bibitem[Choi et~al.(2012)Choi, Wolfe, and Airoldi]{Choi_Wolfe_Airoldi_2012}
D.S. Choi, P.J. Wolfe, and E.M. Airoldi.
\newblock Stochastic blockmodels with a growing number of classes.
\newblock \emph{Biometrika}, 99\penalty0 (2):\penalty0 273--284, Jun. 2012.

\bibitem[Faith et~al.(2007)Faith, Hayete, Thaden, Mogno, Wierzbowski, Cottarel,
  Kasif, Collins, and Gardner]{faith2007clr}
J.J. Faith, B.~Hayete, J.T. Thaden, I.~Mogno, J.~Wierzbowski, G.~Cottarel,
  S.~Kasif, J.J. Collins, and T.S. Gardner.
\newblock Large-scale mapping and validation of escherichia coli
  transcriptional regulation from a compendium of expression profiles.
\newblock \emph{PLoS biology}, 5\penalty0 (1):\penalty0 e8, 2007.

\bibitem[Flynn and Chan(2001)]{flynn2001tb}
J.L. Flynn and J.~Chan.
\newblock Tuberculosis: latency and reactivation.
\newblock \emph{Infection and immunity}, 69\penalty0 (7):\penalty0 4195--4201,
  2001.

\bibitem[Galagan et~al.(2013)Galagan, Minch, Peterson, Lyubetskaya, Azizi,
  Sweet, Gomes, Rustad, Dolganov, Glotova, et~al.]{Galagan2013}
J.E. Galagan, K.~Minch, M.~Peterson, Anna Lyubetskaya, Elham Azizi, Linsday
  Sweet, Antonio Gomes, Tige Rustad, Gregory Dolganov, Irina Glotova, et~al.
\newblock The mycobacterium tuberculosis regulatory network and hypoxia.
\newblock \emph{Nature}, 499\penalty0 (7457):\penalty0 178--183, 2013.

\bibitem[Goldenberg et~al.(2009)Goldenberg, Zheng, Fienberg, and
  Airoldi]{Goldenberg_Zheng_Fienberg_2009}
A.~Goldenberg, A.~X. Zheng, S.~E. Fienberg, and E.~M. Airoldi.
\newblock A survey of statistical network models.
\newblock \emph{Foundations and Trends in Machine Learning}, 2\penalty0
  (2):\penalty0 129--233, Feb. 2009.

\bibitem[Green(1995)]{green1995reversible}
P.J. Green.
\newblock Reversible jump markov chain monte carlo computation and bayesian
  model determination.
\newblock \emph{Biometrika}, 82\penalty0 (4):\penalty0 711--732, 1995.

\bibitem[Imoto et~al.(2003)Imoto, Higuchi, Goto, Tashiro, Kuhara, , and
  Miyano]{Imoto2003a}
S.~Imoto, T.~Higuchi, T.~Goto, K.~Tashiro, S~Kuhara, , and S~Miyano.
\newblock Combining microarrays and biological knowledge for estimating gene
  networks via bayesian networks.
\newblock \emph{Proc. Computational Systems Bioinformatics}, 2003.

\bibitem[Irrthum et~al.(2010)Irrthum, Wehenkel, Geurts,
  et~al.]{irrthum2010genie}
A.~Irrthum, L.~Wehenkel, P.~Geurts, et~al.
\newblock Inferring regulatory networks from expression data using tree-based
  methods.
\newblock \emph{PLoS One}, 5\penalty0 (9):\penalty0 e12776, 2010.

\bibitem[Joshi et~al.(2009)Joshi, De~Smet, Marchal, Van~de Peer, and
  Michoel]{Joshi2009}
A.~Joshi, R.~De~Smet, K.~Marchal, Y.~Van~de Peer, and T.~Michoel.
\newblock Module networks revisited: computational assessment and
  prioritization of model predictions.
\newblock \emph{Bioinformatics}, 25\penalty0 (4):\penalty0 490--496, 2009.

\bibitem[K{\ae}rn et~al.(2005)K{\ae}rn, Elston, Blake, and
  Collins]{kaern2005stochasticity}
M.~K{\ae}rn, T.C. Elston, W.J. Blake, and J.J. Collins.
\newblock Stochasticity in gene expression: from theories to phenotypes.
\newblock \emph{Nature Reviews Genetics}, 6\penalty0 (6):\penalty0 451--464,
  2005.

\bibitem[Kendall et~al.(2007)Kendall, Withers, Soffair, Moreland, Gurcha,
  Sidders, Frita, Ten~Bokum, Besra, Lott, et~al.]{kendall2007highly}
S.L. Kendall, M.~Withers, C.N. Soffair, N.J. Moreland, S.~Gurcha, B.~Sidders,
  R.~Frita, A.~Ten~Bokum, G.S. Besra, J.S. Lott, et~al.
\newblock A highly conserved transcriptional repressor controls a large regulon
  involved in lipid degradation in mycobacterium smegmatis and mycobacterium
  tuberculosis.
\newblock \emph{Molecular microbiology}, 65\penalty0 (3):\penalty0 684--699,
  2007.

\bibitem[Koller and Friedman(2009)]{Koller+Friedman:09}
D.~Koller and N.~Friedman.
\newblock \emph{Probabilistic Graphical Models: Principles and Techniques}.
\newblock MIT Press, 2009.

\bibitem[Kozinets(1999)]{kozinets1999tribalized}
R.V. Kozinets.
\newblock E-tribalized marketing?: The strategic implications of virtual
  communities of consumption.
\newblock \emph{European Management Journal}, 17\penalty0 (3):\penalty0
  252--264, 1999.

\bibitem[Kwak et~al.(2010)Kwak, Lee, Park, and Moon]{Kwak10twittergraph}
H.~Kwak, C.~Lee, H.~Park, and S.~Moon.
\newblock {W}hat is {T}witter, a social network or a news media?
\newblock In \emph{WWW '10: Proceedings of the 19th international conference on
  World wide web}, pages 591--600, New York, NY, USA, 2010. ACM.
\newblock ISBN 978-1-60558-799-8.

\bibitem[Latouche et~al.(2011)Latouche, Birmel{\'e}, and
  Ambroise]{latouche2011overlapping}
P.~Latouche, E.~Birmel{\'e}, and C.~Ambroise.
\newblock Overlapping stochastic block models with application to the french
  political blogosphere.
\newblock \emph{The Annals of Applied Statistics}, 5\penalty0 (1):\penalty0
  309--336, 2011.

\bibitem[Liu et~al.(2013)Liu, Qiao, Zhu, Su, Sun, {Boyd-Kirkup}, and
  Han]{liu2013bayesnetchipseq}
Y.~Liu, N.~Qiao, S.~Zhu, M.~Su, N.~Sun, J.~{Boyd-Kirkup}, and {J.-D.} Han.
\newblock A novel bayesian network inference algorithm for integrative analysis
  of heterogeneous deep sequencing data.
\newblock \emph{Cell Research}, 23\penalty0 (3):\penalty0 440--443, 2013.

\bibitem[Mangan and Alon(2003)]{mangan2003structure}
S.~Mangan and U.~Alon.
\newblock Structure and function of the feed-forward loop network motif.
\newblock \emph{Proceedings of the National Academy of Sciences}, 100\penalty0
  (21):\penalty0 11980--11985, 2003.

\bibitem[Marbach et~al.(2012)Marbach, Costello, K{\"u}ffner, Vega, Prill,
  Camacho, Allison, Kellis, Collins, Stolovitzky, et~al.]{marbach2012wisdom}
D.~Marbach, J.C. Costello, R.~K{\"u}ffner, N.M. Vega, R.J. Prill, D.M. Camacho,
  K.R. Allison, M.~Kellis, J.J. Collins, G.~Stolovitzky, et~al.
\newblock Wisdom of crowds for robust gene network inference.
\newblock \emph{Nature methods}, 2012.

\bibitem[Michoel et~al.(2007)Michoel, Maere, Bonnet, Joshi, Van~den Bulcke,
  Van~Leemput, Van~Remortel, Kuiper, Marchal, et~al.]{michoel2007validating}
T.~Michoel, S.~Maere, E.~Bonnet, Y.~Joshi, A.and~Saeys, T.~Van~den Bulcke,
  K.~Van~Leemput, P.~Van~Remortel, M.~Kuiper, K.~Marchal, et~al.
\newblock Validating module network learning algorithms using simulated data.
\newblock \emph{BMC Bioinformatics}, 8\penalty0 (Suppl 2):\penalty0 S5, 2007.

\bibitem[Mitra et~al.(2013)Mitra, Carvunis, Ramesh, and
  Ideker]{mitra2013integrative}
K.~Mitra, A.~Carvunis, S.K. Ramesh, and T.~Ideker.
\newblock Integrative approaches for finding modular structure in biological
  networks.
\newblock \emph{Nature Reviews Genetics}, 14\penalty0 (10):\penalty0 719--732,
  2013.

\bibitem[Park et~al.(2003)Park, Guinn, Harrell, Liao, Voskuil, Tompa,
  Schoolnik, and Sherman]{park2003dosr}
H.~Park, K.M. Guinn, M.I. Harrell, R.~Liao, M.I. Voskuil, M.~Tompa, G.K.
  Schoolnik, and D.R. Sherman.
\newblock Rv3133c/dosr is a transcription factor that mediates the hypoxic
  response of mycobacterium tuberculosis.
\newblock \emph{Molecular microbiology}, 48\penalty0 (3):\penalty0 833--843,
  2003.

\bibitem[Qi and Ge(2006)]{qi2006modularity}
Y.~Qi and H.~Ge.
\newblock Modularity and dynamics of cellular networks.
\newblock \emph{PLoS Computational Biology}, 2\penalty0 (12):\penalty0 e174,
  2006.

\bibitem[Reddy et~al.(2009)Reddy, Riley, Wymore, Montgomery, DeCaprio, Engels,
  Gellesch, Hubble, Jen, Jin, et~al.]{reddy2009tb}
T.B.K. Reddy, R.~Riley, F.~Wymore, P.~Montgomery, D.~DeCaprio, R.~Engels,
  M.~Gellesch, J.~Hubble, D.~Jen, H.~Jin, et~al.
\newblock Tb database: an integrated platform for tuberculosis research.
\newblock \emph{Nucleic acids research}, 37\penalty0 (suppl 1):\penalty0
  D499--D508, 2009.

\bibitem[Rustad et~al.(2008)Rustad, Harrell, Liao, and
  Sherman]{rustad2008enduring}
T.R. Rustad, M.I. Harrell, R.~Liao, and D.R. Sherman.
\newblock The enduring hypoxic response of mycobacterium tuberculosis.
\newblock \emph{PLoS One}, 3\penalty0 (1):\penalty0 e1502, 2008.

\bibitem[Segal et~al.(2003)Segal, Shapira, Regev, Pe'er, Botstein, Koller, and
  Friedman]{Segal2003}
E.~Segal, M.~Shapira, A.~Regev, D.~Pe'er, D.~Botstein, D.~Koller, and
  N.~Friedman.
\newblock Module networks: identifying regulatory modules and their
  condition-specific regulators from gene expression data.
\newblock \emph{Nature genetics}, 34\penalty0 (2):\penalty0 166--176, 2003.

\bibitem[Segal et~al.(2005)Segal, Pe'er, Regev, Koller, and
  Friedman]{Segal2005}
E.~Segal, D.~Pe'er, A.~Regev, D.~Koller, and N.~Friedman.
\newblock Learning module networks.
\newblock \emph{Journal of Machine Learning Research}, \penalty0 (6):\penalty0
  557--588, 2005.

\bibitem[Snijders and Nowicki(1997)]{snijders1997blockmodel}
T.A.B. Snijders and K.~Nowicki.
\newblock Estimation and prediction for stochastic blockmodels for graphs with
  latent block structure.
\newblock \emph{Journal of Classification}, 14\penalty0 (1):\penalty0 75--100,
  1997.

\bibitem[Voskuil et~al.(2003)Voskuil, Schnappinger, Visconti, Harrell,
  Dolganov, Sherman, and Schoolnik]{voskuil2003inhibition}
M.I. Voskuil, D.~Schnappinger, K.C. Visconti, M.I. Harrell, G.M. Dolganov, D.R.
  Sherman, and G.K. Schoolnik.
\newblock Inhibition of respiration by nitric oxide induces a mycobacterium
  tuberculosis dormancy program.
\newblock \emph{The Journal of experimental medicine}, 198\penalty0
  (5):\penalty0 705--713, 2003.

\bibitem[Wang and Wong(1987)]{wang1987stochastic}
Y.J. Wang and G.Y. Wong.
\newblock Stochastic blockmodels for directed graphs.
\newblock \emph{Journal of the American Statistical Association}, 82\penalty0
  (397):\penalty0 8--19, 1987.

\bibitem[Werhli and Husmeier(2007)]{Werhli2007}
A.V. Werhli and D.~Husmeier.
\newblock Reconstructing gene regulatory networks with bayesian networks by
  combining expression data with multiple sources of prior knowledge.
\newblock \emph{Statistical Applications in Genetics and Molecular Biology},
  6\penalty0 (1):\penalty0 15, 2007.

\bibitem[Yan et~al.(2012)Yan, Xu, and Qi]{yan2012sparse}
F.~Yan, Z.~Xu, and Y.~Qi.
\newblock Sparse matrix-variate gaussian process blockmodels for network
  modeling.
\newblock \emph{arXiv preprint arXiv:1202.3769}, 2012.

\bibitem[Yang and Leskovec(2011)]{yang2011twitter}
J.~Yang and J.~Leskovec.
\newblock Patterns of temporal variation in online media.
\newblock In \emph{Proceedings of the fourth ACM international conference on
  Web search and data mining}, pages 177--186. ACM, 2011.

\bibitem[Yeang and Jaakkola(2006)]{yeang2006modeling}
C.-H. Yeang and T.~Jaakkola.
\newblock Modeling the combinatorial functions of multiple transcription
  factors.
\newblock \emph{Journal of Computational Biology}, 13\penalty0 (2):\penalty0
  463--480, 2006.

\bibitem[Yeang et~al.(2004)Yeang, Ideker, and Jaakkola]{yeang2004physical}
C.-H. Yeang, T.~Ideker, and T.~Jaakkola.
\newblock Physical network models.
\newblock \emph{Journal of computational biology}, 11\penalty0 (2-3):\penalty0
  243--262, 2004.

\end{thebibliography}
%\end{spacing}

%
%
%
%{\begin{figure}[ttt!]
%
%Our model can explain small signal and large signal regulatory effects based on global and condition-specific dependencies. We presented a reversible-jump inference procedure for learning model posterior which can be interpreted based on context.

%\vspace{-3mm}
%0.
%build in causal assumptions in the model to infer and test causal relationships between genes.
%
%
%
%
%\vspace{-2mm}
%\section{Conclusion}
%
%
%
%

\appendix

\setcounter{lem}{0}
\setcounter{thm}{0}
%\setcounter{lem}{0}

% \end{appendices}
\section*{Appendix A. Proof of Lemma \ref{lem1}} \label{theory}
%\vspace{-0.05in}
%Our method uses network data to avoid extra structural assumptions. In this section we formalize this idea through the identifiability of the model we have proposed. This property is important especially for interpretability of learned modules.
%
%Module networks and generally multivariate normal models for node variables can be un-identifiable. In order to obtain an identifiable model based on node variables, imposing extra structural assumptions is necessary. We illustrate that the integrated learning proposed in this paper resolves the un-identifiability issue.
%
%In the following, we first show that modeling node variable alone is identifiable only under very specific conditions. Then, we will restate some results from \cite{latouche2011overlapping} on the identifiability of overlapping block models. Using this result we show the identifiability of the integrated model.
%

\begin{lem}\label{lem1}
{\bf Node Variables Model:} For the model of node variables $\mathbf{X}$, if we have:
\begin{align}
P(\mathbf{X}|\{\mathcal{ A, S}\}',\Theta', \Sigma')=P(\mathbf{X}|\{\mathcal{ A, S}\},\Theta, \Sigma)
\end{align}
\begin{enumerate}
\item Then, we can conclude: $\mu'=\mu \ and \ \Sigma'=\Sigma$.
%\vspace{-2mm}
\item If we further assume $\{\mathcal{ A, S}\}=\{\mathcal{ A, S}\}'$ and that each module has at least two non parent nodes and $\sum_k |Pa_k| < N$ and the covariance matrix $\Sigma$ is invertible, we can conclude: $\Theta=\Theta'$, $W=W'$.
\end{enumerate}
\end{lem}

\begin{sketch}
\begin{enumerate}
\item Considering that distributions of $\mathbf{X}$ are multivariate Normal under both parameter sets, it is straight forward that the mean and covariance parameters of two Normals should be the same. This can be formally shown by finding maximum of the distribution and curvature at any point for both sides, hence, $\mu'=\mu$ and $\Sigma'=\Sigma$.
\item From the identifiability of $\mu$ and $\Sigma$, it is sufficient to show that $\mu$ and $\Sigma$ uniquely define $\Theta$, $W$ given $\{\mathcal{ A, S}\}$. Starting from $\Gamma_c$, we can consider the following set of linear equations:
    \begin{align*}
    {\boldsymbol \mu_c}=  {\Gamma}_c {\boldsymbol \mu_{c}^R}
    \end{align*}
    This is a set of equations with $N$ equations and $\sum_k |Pa_k|$ unknowns. Hence, when $\sum_k |Pa_k| < N$ this set of linear equations will lead to a unique solution if a solution exists.

    For the $\Sigma$, given that it is invertible, we have:
    \begin{align}
    \Sigma^{-1}=(I-W)^{T}(I-W)
    \end{align}
    Considering that parents have the same value for $W_{nr}$ for $\forall n \in M_k$. Then, we can simply find $W_{nr}$ by solving $|Pa_k|*{W_{nr}}^2=\Sigma^{-1}_{ij}$ where $i,j$ are two genes that are non parents and belong to the module $M_k$.
\end{enumerate}
\end{sketch}
%

%The above lemma provides identifiability for the case where the structure $\{\mathcal{ A, S}\}$ is assumed to be known. However, in the case that we don't have the structure, the parameterizations of multivariate normal ($\mu$ and $\Sigma$) can be written in multiple ways in terms of $\Theta$ and $\{\mathcal{ A, S}\}$. This is due to existence of multiple decompositions for the covariance matrix.

\section*{Appendix B. Proof of Theorem \ref{thm1}}

%In following, we will use a theorem for identifiability of overlapping block models from~\cite{latouche2011overlapping} which is an extension of the results in~\cite{allman2009identifiability}. The results provide conditions for overlapping stochastic block models to be identifiable.
%%
\begin{thm}\label{thm1}
{\bf Network data Model:} If we have:
\begin{align}
P(\mathbf{B}| \{{\mathcal A,S}\},\boldsymbol{\pi})&=P(\mathbf{B}| \{{\mathcal A,S}\}',\boldsymbol{\pi}')
\end{align}
Then: $\{{\mathcal A,S}\}=\{{\mathcal A,S}\}'$ with a permutation and $\boldsymbol{\pi}=\boldsymbol{\pi}'$(except in a set of parameters which
have a null Lebesgue measure).
\end{thm}

\begin{sketch}
Our network data model is an overlapping stochastic block model, where the blocks are parents and modules, with a specific parametrization among the modules and parents. Hence, we have the identifiability using the Theorem 4.1 in ~\cite{latouche2011overlapping}.
\end{sketch}

\section*{Appendix C. Proof of Theorem \ref{thm2}}

%Using the above Theorem and Lemma \ref{lem1} we can have the following Theorem for the identifiability of the integrated model.
%%
\begin{thm}
{\bf Identifiability of model:} If we have:
\begin{align}
P(\mathbf{B}| \{{\mathcal A,S}\},\boldsymbol{\pi})&=P(\mathbf{B}| \{{\mathcal A,S}\}',\boldsymbol{\pi}')\label{relation}\\
P(\mathbf{X}|\{\mathcal{ A, S}\}',\Theta', \Sigma')&=P(\mathbf{X}|\{\mathcal{ A, S}\},\Theta, \Sigma)\label{variable}
\end{align}
with assuming that each module has at least two non-parent nodes and $\sum_k |Pa_k| < N$ and the covariance matrix $\Sigma$ is invertible, then: $\{{\mathcal A,S}\}=\{{\mathcal A,S}\}'$ with a permutation, $\boldsymbol{\pi}=\boldsymbol{\pi}'$ , $\Theta=\Theta'$ and $W=W'$.
\end{thm}

\begin{sketch}
This theorem is an immediate result from combination of Theorem \ref{thm1} and Lemma \ref{lem1}.
Using (\ref{relation}), according to Theorem \ref{thm1} we have: $\{{\mathcal A,S}\}=\{{\mathcal A,S}\}'$ with a permutation and $\boldsymbol{\pi}=\boldsymbol{\pi}'$. Now, knowing $\{{\mathcal A,S}\}=\{{\mathcal A,S}\}'$ and equation (\ref{variable}) we can apply Lemma \ref{lem1} leading to $\Theta=\Theta'$ and $W=W'$. This concludes the proof.
\end{sketch}

\section*{Appendix D. Learning Module Assignment $\mathcal{A}$.} \label{sec:apxe}
Learning the assignment of each gene to a module, involves learning the number of modules. Changing the number of modules however, changes dimensions of the parameter space and therefore, densities will not be comparable. Thus, to sample from $P(\mathcal{ A}|\mathcal{S},\Theta, \Sigma,,Z^S \boldsymbol{\pi},\mathbf{X},\mathbf{B})$, we use the Reversible-Jump MCMC method \citep{green1995reversible}, an extension of the Metropolis-Hastings algorithm that allows moves between models with different dimensionality.
%$$\vspace{-0.8in}$$

%\vspace{-3mm}
%\vspace{-12mm}
%\end{wrapfigure}

%
%
%
In each proposal, we consider three close move schemes
%on assignment function $\mathcal{A}$
of increasing or decreasing the number of modules by one, or not changing the total number. For increasing the number of modules, a random gene is moved to a new module of its own and for decreasing the number, two modules are merged. In the third case, an gene is randomly moved from one module to another module, to sample its assignment.

We design transformation of parameters using Green's method to extend model dimensions (Algorithm \ref{alg2}) The acceptance ratio for the split move is $P_{split} = \min \{1, \frac{P(\mathcal{M}^{(i+1)}|X,B)}{P(\mathcal{M}^{(i)}|X,B)} \times \frac{\frac{1}{K+1}}{\frac{1}{K}}  \times \frac{p_{+1}}{p_{-1}} \times  \frac{1}{p({\bf{u}})p({\bf{u'}})} \times \mathcal{J}_{(i) \rightarrow (i+1)} \}$
where $\mathcal{J}_{(i) \rightarrow (i+1)}$ is the Jacobian of the transformation from the previous state to the proposed state, and the acceptance ratio for the merge move is $P_{merge} = \min \{1, \frac{P(\mathcal{M}^{(i+1)}|X,B)}{P(\mathcal{M}^{(i)}|X,B)} \times \frac{\frac{1}{K-1}}{\frac{1}{K}} \times \frac{p_{-1}}{p_{+1}} \times \mathcal{J}_{(i) \rightarrow (i+1)}  \}$.

    \begin{minipage}[t]{1\textwidth}
   % \vspace{-6mm}
    \flushleft
\begin{algorithm}[H]
   \caption{RJMCMC to update $\mathcal{A}$} % from ${\mathcal{A}}^{(i)}$}
   \label{alg2}
\begin{algorithmic}[1]
%\STATE {\bfseries Inputs:} $\mathbf{X},\mathbf{B}$; $\mathcal{A}^{(i)},\mathcal{S}^{(i)},\Theta^{(i)}, \Sigma^{(i)}, \boldsymbol{\pi}^{(i)}$
   %\STATE Set $p_{-1},p_{0},p_{+1}$
   \STATE Find $K$: number of distinct modules in $\mathcal{A}^{(i)}$
   \STATE Propose move $\nu$ from $\{-1,0,+1\}$ with probabilities $p_{-1},p_{0},p_{+1}$, respectively.
   \SWITCH {$\nu$}
   \CASE {+1}
   %\STATE Select random module $M_k$ uniformly
   \STATE Select random gene $n \in M_k$ uniformly
   \STATE Assign $n$ to new module $M_{K+1}$ %and remainder to $M_{k_2}$
   \STATE Assign parents $Pa_{K+1} =Pa_{k} $
   \STATE Draw vectors $ {\bf u},{\bf u'} \sim \mathcal{N}(0,1)$
   \STATE Propose parameters:
   \STATE $\boldsymbol{\pi}_{k1}^{Pa_{K+1}}=\boldsymbol{\pi}_{k}^{Pa_{k}} - {\bf u}$, $\boldsymbol{\pi}_{k2}^{Pa_{k}}=\boldsymbol{\pi}_{k}^{Pa_{k}} + {\bf u}$
   %\STATE Draw vector $ {\bf u'}\sim \mathcal{N}(0,1)$
   \STATE $\boldsymbol{\gamma}_{k1}^{Pa_{K+1}}=\boldsymbol{\gamma}_{k}^{Pa_{k}} - {\bf u'}$, $\boldsymbol{\gamma}_{k2}^{Pa_{k}}=\boldsymbol{\gamma}_{k}^{Pa_{k}} + {\bf u'}$
   \STATE Compute $\{\Theta,\Sigma, \boldsymbol{\pi}\}$
   \STATE Accept $\mathcal{A}^{(i+1)}$ with $P_{split}$ %from (\ref{split})
   %\ELSE
   \ENDCASE
   \CASE {$-1$}
   \STATE Select two random modules $M_{k_1}$ and $M_{k_2}$
   \STATE Merge into one module $M_k1$
   \STATE Assign parents $Pa_{k1}=Pa_{k_1} \cup Pa_{k_2}$
   \FOR{$\forall r \in Pa_{k_1} \cap Pa_{k_2}$}
   \STATE Propose $\boldsymbol{\pi}_{k1}^{r}=(\boldsymbol{\pi}_{k1}^{r}+\boldsymbol{\pi}_{k2}^{r})/2$
   \STATE and   $\boldsymbol{\gamma}_{k1}^{r}=(\boldsymbol{\gamma}_{k1}^{r}+\boldsymbol{\gamma}_{k2}^{r})/2$
   \ENDFOR
   \STATE Compute $\{\Theta,\Sigma, \boldsymbol{\pi}\}$
   \STATE Accept $\mathcal{A}^{(i+1)}$ with $P_{merge}$ %from (\ref{merge})
   \ENDCASE
   \CASE {0}
   \STATE Select two random modules $M_{k_1}$, $M_{k_2}$
   \STATE Move a random gene $n$ from $M_{k_1}$ to $M_{k_2}$
   \STATE Compute $\{\Theta,\Sigma, \boldsymbol{\pi}\}$
   \STATE Accept $\mathcal{A}^{(i+1)}(n)=k_2$ with $P_{mh}$ %from (\ref{mh})
   %\ENDIF
   %\ENDIF
   \ENDCASE
   \ENDSWITCH
   %\STATE
   %\STATE {\bfseries return} {$\mathcal{A}^{(i+1)}$}
\end{algorithmic}
\end{algorithm}
%\vspace{2mm}
\end{minipage}

%Upon acceptance of the proposal in each of the three cases, the corresponding dependant parameters are also updated based on assignment.
%\begin{align}\label{merge}
%    P_{merge} = \min \{1, \frac{P(\mathcal{M}^{(i+1)}|X,B)}{\mathcal{L}(\{ \mathcal{M}^{(i)};X,B)} \times \frac{\frac{1}{K-1}}{\frac{1}{K}} \times \nonumber \\ \frac{p_{-1}}{p_{+1}} \times \mathcal{J}_{(i) \rightarrow (i+1)}  \}
%\end{align}
%\begin{equation}\label{merge}
%    P_{merge} = \min \{1, \frac{P(\mathcal{M}^{(i+1)}|X,B)}{\mathcal{L}(\{ \mathcal{M}^{(i)};X,B)} \times \frac{\frac{1}{K-1}}{\frac{1}{K}} \times \frac{p_{-1}}{p_{+1}} \times \mathcal{J}_{(i) \rightarrow (i+1)}  \}
%\end{equation}
%\vspace{-2mm}
%
%
%
%
%
%\begin{wrapfigure}{}{0.48\textwidth}
%\vspace{-3mm}
%\end{wrapfigure}
%\begin{wrapfigure}{r}{0.45\textwidth}
%$$\vspace{-0.5in}$$
%\vspace{-3mm}
%
\section*{Appendix E. Learning Dependency Structure $\mathcal{S}$.}
To sample from the dependency structure $P(\mathcal{ S}|\mathcal{A},\Theta, \Sigma,Z^S \boldsymbol{\pi},\mathbf{X},\mathbf{B})$ (assignment of parents), we also implement a Reversible-Jump method, as the number of parents for each module needs to be determined. Two proposal moves are considered for $\mathcal{S}$ which include increasing or decreasing the number of parents for each module, by one (Algorithm \ref{alg3}).
%
%$P(\mathcal{S}^{(i+1)}|\mathcal{A}^{(i),\mathcal{S}^{(i)},\Theta^{(i)}, \Sigma^{(i)}, \boldsymbol{\pi^{(i)}},\mathbf{X},\mathbf{B})$}
%
%
%Dependent parameters are then updated accordingly.
In the case of addition of a parent to a module, we propose mixture coefficients $\gamma$ and interaction parameters $\pi$ for the added regulator, based on its learned values in another module, where it has already been assigned as a parent, with an additional noise term. The acceptance ratio for the add proposal is $ P_{add} = \min \{1, \frac{P(\mathcal{M}^{(i+1)}|X,B)}{P(\mathcal{M}^{(i)}|X,B)} \times   \frac{\frac{1}{R_k+1}}{\frac{1}{R-R_k}} \times \frac{p_{S}}{1-p_{S}} \times \frac{1}{p({{u}})p({\bf{u'}})} \times \mathcal{J}_{(i) \rightarrow (i+1)} \}$
%
%\begin{align}\label{add}
%    P_{add} = \min \{1, \frac{P(\mathcal{M}^{(i+1)}|X,B)}{P(\mathcal{M}^{(i)}|X,B)} \times   \frac{\frac{1}{R_k+1}}{\frac{1}{R-R_k}} \times \frac{p_{S}}{1-p_{S}} \times \frac{1}{p({{u}})p({\bf{u'}})} \times \mathcal{J}_{(i) \rightarrow (i+1)} \}
%\end{align}
%
where $R_k$ is the number of parents for module $k$ in the $i-$th state,
and the acceptance ratio for the remove proposal is $    P_{rem} = \min \{1, \frac{P(\mathcal{M}^{(i+1)}|X,B)}{P(\mathcal{M}^{(i)}|X,B)} \times  \frac{\frac{1}{R-R_k+1}}{\frac{1}{R_k}} %\nonumber
    \times \frac{1-p_{S}}{p_{S}} \times \mathcal{J}_{(i) \rightarrow (i+1)} \}$.

\begin{minipage}{1\textwidth}
  % \vspace{-6mm}
\begin{algorithm}[H]
   \caption{RJMCMC to update $\mathcal{S}$} %from $\mathcal{S}^{(i)}$}
   \label{alg3}
\begin{algorithmic}[1]
%\STATE {\bfseries Inputs:} $\mathbf{X},\mathbf{B}$; $\mathcal{A}^{(i)},\mathcal{S}^{(i)},\Theta^{(i)}, \Sigma^{(i)}, \boldsymbol{\pi}^{(i)}$
   \STATE Set $p_S$
   \FOR{module $k=1$  {\bfseries to} $K$}
   \STATE Propose $\nu$ from $\{+1,-1\}$  with $p_S$
   \SWITCH {$\nu$}
   \CASE {$+1$}
   \STATE Add a random parent $r \in {1,...,R}$ to $Pa_k$
   \STATE Draw $u,{\bf u'} \sim Unif(0,1)$
   \IF{$r$ is also a parent of another module $Pa_{k'}$}
   \STATE Propose $\pi_k^r = \pi_{k'}^r + u$, $\gamma_c^{r_{k}}=\gamma_c^{r_{k'}}+{\bf u'}(c)$ for all $c \in \{1,...,C\}$
   \ELSE
   \STATE Propose $\pi_k^r = u$,$\gamma_c^{r_{k}}={\bf u'}(c)$ for all $c$
   \ENDIF
   \STATE Compute $\{\Theta,\Sigma, \boldsymbol{\pi}\}$
   \STATE Accept $\mathcal{S}^{(i+1)}$ with $P_{add}$% from equation (\ref{add})
   \ENDCASE
   \CASE {$-1$}
   \STATE Remove a random parent $r$ from $Pa_k$
   \STATE Compute $\{\Theta,\Sigma, \boldsymbol{\pi}\}$
   \STATE Accept $\mathcal{S}^{(i+1)}$ with $P_{rem}$ % from equation (\ref{remove})
   \ENDCASE
   \ENDSWITCH
   \ENDFOR
%\STATE {\bfseries return} $\mathcal{S}^{(i+1)}=\{Pa_1^{(i+1)},...,Pa_K^{(i+1)} \}$
\end{algorithmic}
\end{algorithm}
\end{minipage}

\end{document}